\documentclass[sn-basic,iicol]{sn-jnl}

\jyear{2021}%

\theoremstyle{thmstyleone}%
\theoremstyle{thmstyletwo}%

\theoremstyle{thmstylethree}%

\usepackage[utf8]{inputenc}
\usepackage{afterpage}
\usepackage{graphicx}
\usepackage{lmodern,babel,adjustbox,booktabs,multirow}
\usepackage{booktabs, siunitx}
\usepackage{comment}
\usepackage[para,online,flushleft]{threeparttable}
\usepackage{tabularx}
\usepackage{longtable}
\usepackage{adjustbox}
\usepackage[center]{caption}
\usepackage{lipsum}
\begin{document}

\title[Article Title]{A comprehensive survey on recent deep learning-based methods applied to surgical data}


\author*[1]{\fnm{Mansoor} \sur{Ali}}\email{a01753093@tec.mx}

\author[1]{\fnm{Rafael Martínez} \sur{García Peña}}\email{a01274853@tec.mx}
\author[1]{\fnm{Gilberto Ochoa} \sur{Ruiz}}\email{gilberto.ochoa@tec.mx}
\author[2]{\fnm{Sharib} \sur{Ali}}\email{s.s.ali@leeds.ac.uk}


\affil*[1]{\orgdiv{Escuela de Ingeniería y Ciencias}, \orgname{Tecnologico de Monterrey}, \orgaddress{\street{Av. Eugenio Garza Sada}, \city{Monterrey}, \postcode{38115}, \state{Nuevo Leon}, \country{Mexico}}}

\affil[2]{\orgdiv{School of Computing}, \orgname{University of Leeds}, \orgaddress{\city{Leeds} \country{UK}}}



\abstract{Minimally invasive surgery is highly operator dependant with a lengthy procedural time causing fatigue to surgeon and risks to patients such as injury to organs, infection, bleeding, and complications of anesthesia. To mitigate such risks, real-time systems are desired to be developed that can provide intra-operative guidance to surgeons. For example, an automated system for tool localization, tool (or tissue) tracking, and depth estimation can enable a clear understanding of surgical scenes preventing miscalculations during surgical procedures. In this work, we present a systematic review of recent machine learning-based approaches including surgical tool localization, segmentation, tracking, and 3D scene perception. Furthermore, we provide a detailed overview of publicly available benchmark datasets widely used for surgical navigation tasks. While recent deep learning architectures have shown promising results, there are still several open research problems such as a lack of annotated datasets, the presence of artifacts in surgical scenes, and non-textured surfaces that hinder 3D reconstruction of the anatomical structures. Based on our comprehensive review, we present a discussion on current gaps and needed steps to improve the adaptation of technology in surgery.} 





 \maketitle


\begin{sloppypar}

\section{Introduction}
%
\begin{sloppypar}

Over time surgeons have successively  updated  surgical intervention techniques by assessing underlying deficiencies and their subsequent impact on patient hospitalization and recovery times \citep{3}. Open surgery is usually performed to treat various diseases when lesions of interest lie inside the human body. However, this approach can cause several inconveniences to patients, leading to longer recovery times, increased susceptibility to hospital-related infections and long-term scars. Alternatively, minimally invasive surgery (MIS), which was developed in the 1990s has been gaining traction in recent years. This approach employs smaller incisions to introduce a  surgical instrument, along with a camera-mounted endoscope to aid in the exploration of the internal organs of the patient. Over the past decade, the medical field has witnessed an exponential rise in the adoption of MIS as a preferred choice to ensure patient safety, with an adoption rate beyond 80\% of the overall cases \citep{tsui2013:springer}. 
MIS was introduced with the promise of several therapeutic benefits for the patients such as reduced trauma, lesser risks of post-operative complications, potentially enhanced safety, quicker recovery times, and faster hospital discharge. Furthermore, the endoscope used during MIS procedures provide a faster and more effortless way to capture images and record videos of the surgical procedure \citep{12}.


However, the benefits of MIS come at the cost of an increased complexity compared to open surgical procedures. During MIS procedures, surgeons have to perform the surgical procedure indirectly through a monitor that displays the video signal from an endoscopic camera. Therefore, MIS may be quite challenging for the operator on two counts: first, to maintain hand-eye coordination throughout the procedure and secondly, keeping a safe distance between the instruments and the surrounding sensitive tissues. MIS is therefore considered as a highly operator dependant procedure and thus there is a need for assistive tools and technologies that can ensure the patients' safety and surgeons' confidence in these procedures. 


In this paper, we argue that the AI-related developments in the following three directions can help alleviate the above-mentioned challenges. These include:

\begin{enumerate}
    \item Surgical tool localization 
    \item Surgical procedural analysis
    \item Surgical scene understanding
\end{enumerate}

\begin{figure*}[h!]
\centering
\includegraphics[width=6.2in]{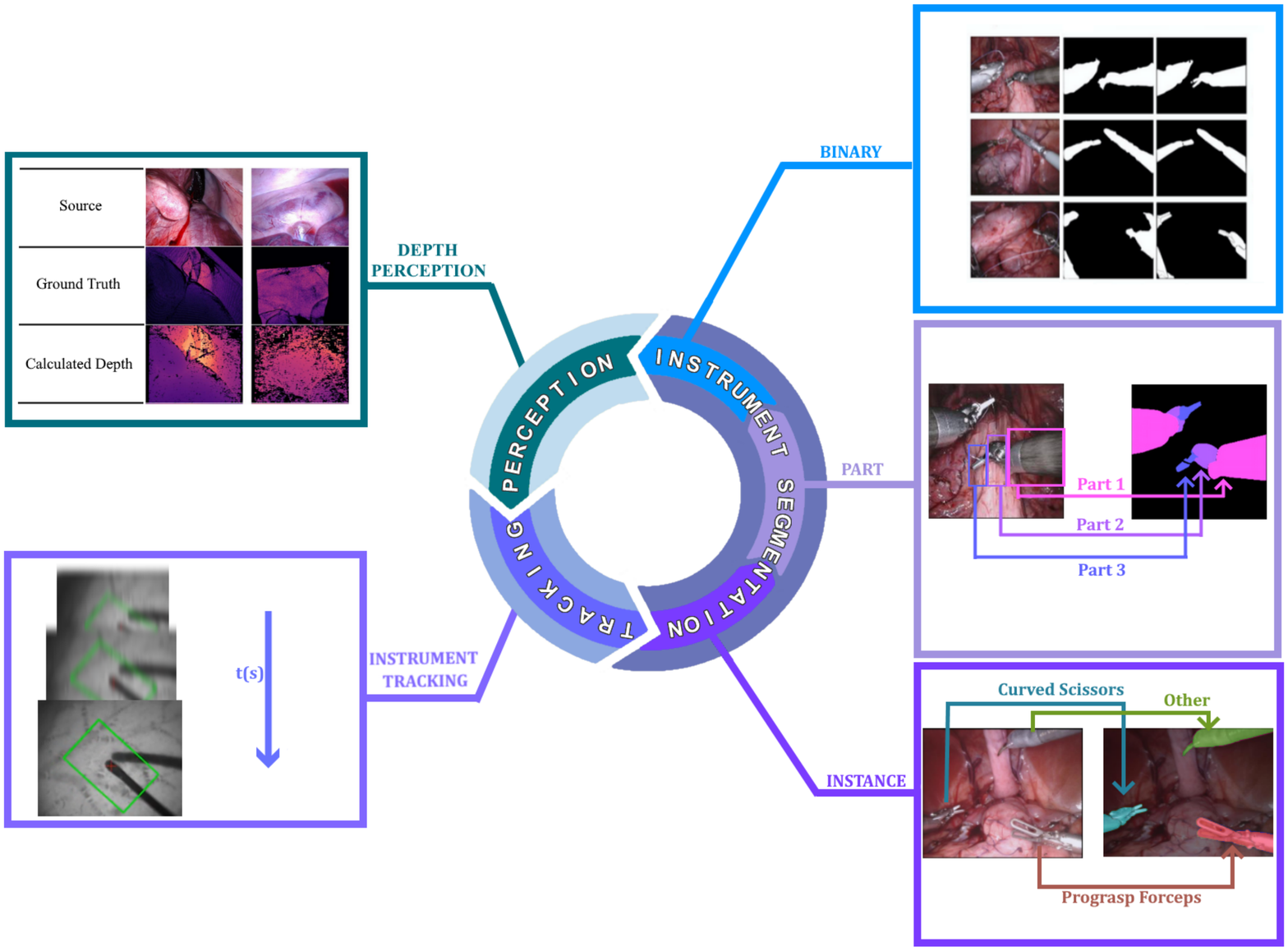}
\caption{Overview of surgical tool navigation tasks}
\label{Overview}
\end{figure*}

Surgical navigation and scene depth information can provide a much-needed aid for the surgical team for a better scene understanding and information processing. Recently, computer vision and deep learning (DL) have achieved state-of-the-art (SOTA) performance in image classification \citep{ciregan2012multi}, object detection \citep{zhao2019object}, segmentation \citep{long:CVPR}, tracking \citep{wang2016stct} and 3D scene reconstruction \citep{yang2021dense:MIUA}. In this context, surgical navigation, and depth perception have been investigated by researchers in terms of tool segmentation, detection, tracking, and 3D scene reconstruction. In this paper, we present a comprehensive review of these tasks tailored for surgical assistance. Fig.\ref{Overview} shows the general overview of the surgical navigation tasks.




%

%

Surgical data science has shown a great deal of progress due to the efforts done by groups working on surgical data curation, as well as the development of novel technologies to assist surgical procedures that have accelerated this field \citep{camma,ucl}.
Improved access to diverse surgical data has fostered the development of efficient DL techniques for various clinical assistance tasks. 
Surgical data science (SDS) was defined as a scientific discipline ``to acquire, organize, analyze, and model the data to improve the quality of interventional healthcare'' in a workshop held in Heidelberg Germany in 2016~\citep{9}. SDS can provide great assistance during the entire clinical pathway of a patient from decision support to context-aware assistance and surgical training~\citep{10}. MIS is one of the most important applications of SDS. Herein, data gathered through the endoscope can be of immense assistance in numerous surgical navigation applications like surgical tool detection \citep{13,14}, segmentation \citep{16,17}and tracking \citep{19,20}, scene depth estimation and 3D reconstruction \citep{21,22}, surgeon's skill assessment \citep{23}, gesture recognition \citep{24}, workflow modelling \citep{25}, scene segmentation \citep{jin2022exploring} and surgical report generation \citep{xu2021class:MICCAI}.   



Several research works have been conducted recently on surgical navigation and depth perception using DL-based methods.  Despite these great strides, several research gaps still persist. Broadly speaking, the lack of labeled data, the unavailability of diverse multi-center datasets, and the real-time usability  along with model robustness and generalizability capabilities are still open problems in computer vision for MIS. The lack of labeled data has been approached through different techniques such as transfer learning, meta-learning, semi and weakly supervised and learning or using synthetic data. Synthetic data in addition to various augmentation strategies have been used to tackle model robustness while adversarial approaches are being used to improve model generalizability. 

In this work, we provide a detailed survey of trends in DL-based techniques used in the surgical tool navigation tasks, specifically tool detection and classification, tool instance segmentation, tracking and surgical scene depth perception.

\begin{figure*}[t!]
\centering
\includegraphics[width=6.3in]{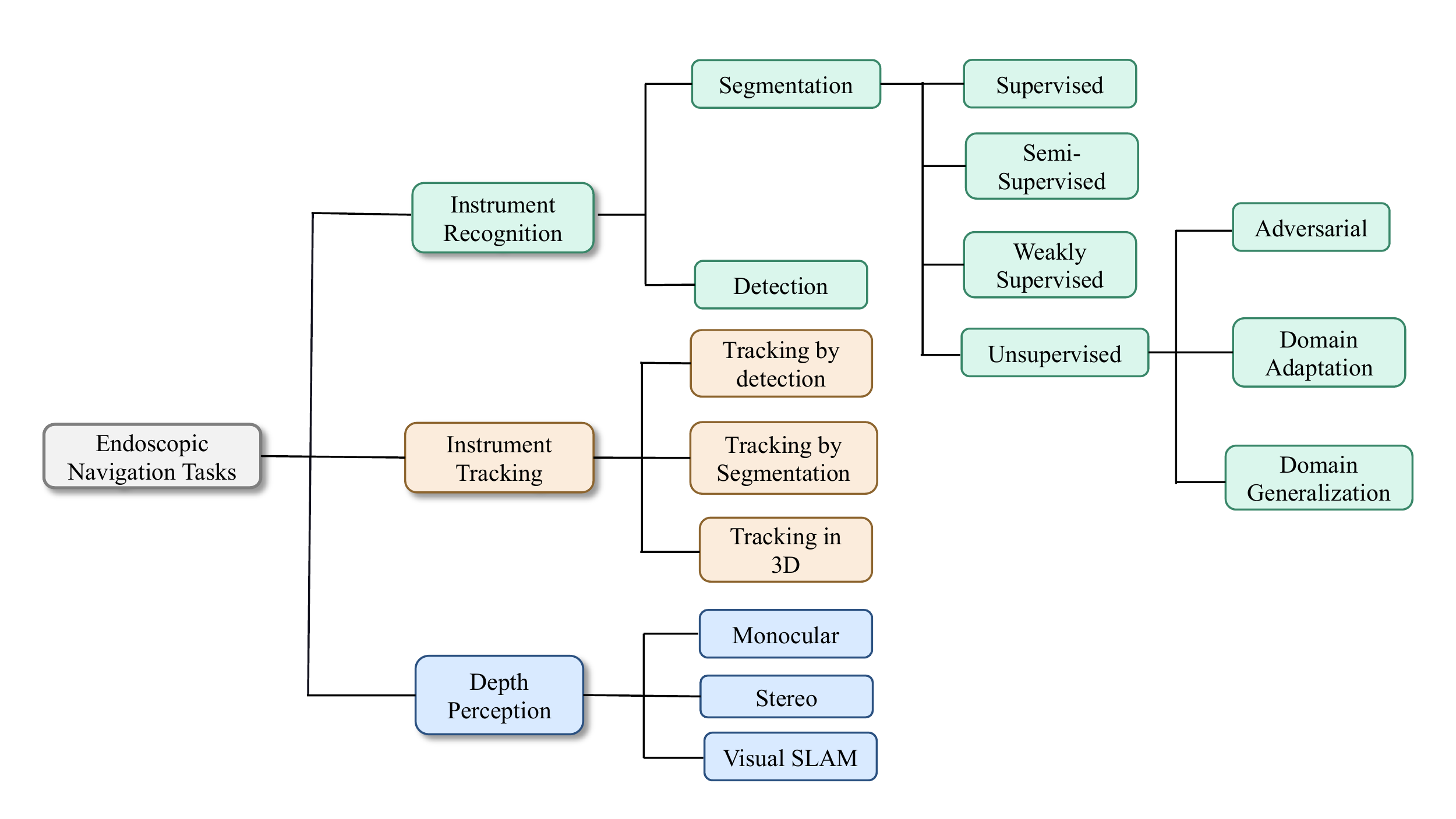}
\caption{Categorization of surgical tool navigation tasks and approaches}
\label{fig2}
\end{figure*}

The rest of this paper is organized as follows. In Section \ref{sec2}, we discuss the criteria for the selection of papers used in this study. Section \ref{sec3} highlights previously written reviews on the relevant aspects of this work and objectively compares them with this work. Section \ref{sec4} outlines challenges in surgical tool navigation. How these challenges have been tackled is the topic of the next section. Section \ref{sec5} of the paper is divided into different sections as depicted in Fig.\ref{fig2}. It provides a comprehensive review of SOTA in instrument segmentation, detection, tracking, and depth perception. Section \ref{sec6} presents the overview of public datasets, current research gaps and future directions are provided in Section \ref{sec7} and finally conclusion is provided in Section \ref{sec8}. 


%

\section{Selection of papers}\label{sec2}
The aim of this survey article is to analyze and critically assess the current SOTA in surgical navigation, more specifically, detection, segmentation, tracking of surgical instruments, and depth perception of surgical scenes. For this review, we have prioritized top journal and conference papers.  A thorough search was conducted on Google Scholar, PubMed, arXiv, Springer, Elsevier, ACM Digital Library and IEEE Xplore to find pertinent works related to surgical navigation. The search results were filtered for the past five years since our study only covers DL-based architectures. For a detailed account of pre-DL methods, the interested reader is directed to excellent papers such as \citep{saint2011,blum2010modeling}. For other orthogonal areas of MIS such as  gesture recognition, workflow modeling, scene segmentation and surgical report generation, following sources can be consulted \citep{24,25,jin2022exploring,xu2021class:MICCAI,saint2011,blum2010modeling}. We used the following search terms: "Endoscopic navigation" OR "Computer-Assisted Surgery" OR "Surgical tool Segmentation" OR "MIS" AND "Instrument Detection" OR "Depth Perception" OR "Endoscopic instrument tracking" OR "Three dimensional endoscopy".


An initial screening based on the paper title was performed. During this process, duplicates, summaries, abstracts, doctoral symposiums, tutorials, book entries, and survey works were discarded. Afterward, papers were screened based on the following criteria: 
\begin{itemize}
 
 \item Works based on the type of input data used. We focused on computer vision-based methods for this study, excluding the works that use kinematic data or any other modality. 
 \item Papers based on traditional computer vision or ML algorithms were excluded from this study. Instead, we focused only on recent deep learning-based methods for this survey. 
 
 \item This work mainly focuses on surgical navigation tasks such as tool segmentation, tracking and depth perception. Thus, to make the study more precise, papers not found to be relevant to the scope of this study were removed from the detailed analysis. 
\end{itemize}

This screening process finally left us with 100 papers for the comprehensive review. A further breakdown of the paper selection is described as follows: 46 papers for instrument segmentation, 21 for instrument bounding box detection and recognition, 13 papers for instrument tracking and 20 for depth perception.

\begin{table}[h]
\tabcolsep7.5pt
\caption{Acronyms used throughout the paper}\label{tab1}%
\begin{center}
\begin{tabular}{@{}ll@{}}
\hline
MTA    & Multimedia Tools and Applications \\
MIA    & Medical Image Analysis  \\
CAS    & Computer Assisted Surgery  \\
IJCARS & International Journal of Computer \\
& Assisted Radiology and Surgery\\
IJMRCAS  & International Journal of Medical  \\
&Robotics\\
& and Computer Assisted Surgery\\
CVAH  & Computer Vision for Assistive \\
&Healthcare\\
FH  & Frontiers of Medicine\\
IEEE RBM  & IEEE Reviews in Biomedical \\
&Engineering\\
SE & Surgical Endoscopy\\
CNN & Convolutional Neural Networks\\
MIS & Minimally Invasive Surgery\\
SDS & Surgical Data Science\\
VSLAM & Visual simultaneous localization and \\
&mapping\\
I2I & Image-to-Image\\
SOTA& State-of-the-art\\
\hline
\end{tabular}
\end{center}
\end{table}

\section{Comparison with previous relevant reviews}\label{sec3}
Various survey articles have been published on surgical navigation tasks such as instrument segmentation, tracking, and depth perception in the literature. A summary of previous reviews is provided in Table \ref{tab2}. Some of these surveys cover MIS extensively. For instance, authors in \cite{munzer2018content} and \cite{zhou2020application} provide detailed analysis of MIS procedures by breaking them down into three separate aspects: pre-processing, real-time support at procedure time and post-operative phases. However, both surveys discuss the SOTA methods very briefly. 


The paper by \cite{sorensen2016three} outlines the effect of having a three-dimensional internal view of the procedure on the screen as compared to a 2D view. However, it does not delve into a thorough discussion of recent techniques; the paper mostly covers aspects such as the impact of 3D visualization on performance time, precision errors and cognitive load. 
The study in \cite{bodenstedt2018comparative} provide a comparative analysis of DL-based architectures with traditional techniques for instrument segmentation and tracking. Most of the methods reviewed in this work are either based on CNN  or random forests machine learning models. The major contribution of the paper is that validation data is generated to test the applicability of participating methods.\cite{yang2020image} provide a  comprehensive overview of instrument segmentation and tracking methods based partially or fully on CNN methods. An overview of the recent methods for instrument tracking in the field of retinal microsurgery is discussed in \cite{rieke2018computer}. A review of 3D Reconstruction methods, feature detection and tracking methods is presented in \cite{lin2016video}, while a performance analysis of 3D computer vision in laparoscopic surgery, as compared to 2D is the focus of study in \cite{sorensen2016three}. In a more recent survey, \cite{rivas2021review:Acess} reviews the use of DL methods for various tasks in MIS such as  tool detection, segmentation, phase recognition and skill assessment. Though the work covers several aspects of MIS, the analysis related to tool segmentation is restricted to supervised and semi-supervised methods.
\begin{table*}[!htbp]
\begin{minipage}{230pt}
\tabcolsep7.5pt
\setlength{\tabcolsep}{8pt}
\centering
\caption{Summary of previous relevant reviews}\label{tab2}
\begin{tabular}{@{} |p{0.05\linewidth} | p{0.63\linewidth} | p{0.17\linewidth}| p{0.18\linewidth} | p{0.7\linewidth}|}
\hline
\textbf{No.} & \textbf{Survey Title}  & \textbf{Ref.} & \textbf{Published} & \textbf{Content} \\
\hline
1    & Surgical robotics beyond enhanced dexterity instrumentation: a survey of machine learning techniques and their role in intelligent and autonomous surgical actions   & \cite{kassahun2016surgical}  & IJCARS & Role of ML in the context of surgery with focus on Surgical robotics \\
\hline
2   & Video-based 3D reconstruction, laparoscopic localization and deformation recovery for abdominal minimally invasive surgery: a survey  & \cite{lin2016video}  &  IJMRCAS & Review of SOTA MIS-VSLAM techniques for abdominal MIS \\
\hline
3    & Three-dimensional versus two-dimensional vision in laparoscopy:
a systematic review   & \cite{sorensen2016three} & SE & Effect of 3D vision on laproscopic performance compared to 2D Laproscopy\\
\hline
4    & Vision-based and marker-less surgical tool detection and tracking: a review of the literature   & \cite{bouget2017vision}  & MIA & Discuss Tool detection and tracking in terms of Validation Data-sets, Methodology and Detection Methods \\
\hline
5    & Content-based processing and analysis of endoscopic images and videos: A survey   & \cite{munzer2018content}  & MTA & Holisitc endoscopic video analysis in pre-processing, intr-operative and post processing phases \\
\hline
6   & Comparative evaluation of instrument segmentation and tracking methods in minimally invasive surgery   & \cite{bodenstedt2018comparative} & arXiv & Comparative Summary of segmentation and tracking methods from EndoVis 2015 Challenge \\
\hline
7    & Computer Vision and Machine Learning for Surgical Instrument Tracking \footnotemark[1]  & \cite{rieke2018computer}   & CVAH & Challenges and requirements for retinal microsurgery instrument tracking and their performance evaluation \\
\hline
8   & Optical and electromagnetic tracking systems for biomedical applications: a critical review on potentialities and limitations
 & \cite{sorriento2019optical}   & IEEE RBM & Comparison of Electromagnetic and optical tracking systems for Biomedical applications \\
\hline
9    & Image-based laparoscopic tool detection and tracking using convolutional neural networks: a review of the literature   & \cite{yang2020image}   & CAS & Partial and Full CNN based approaches for tool detection and tracking  \\
\hline
10   & Application of artificial intelligence in surgery
 & \cite{zhou2020application}   & FM & Review of AI applications in endoscopic pre-operative, itra-operative phases and use in surgical robotics \\
\hline
11    & The Future of Endoscopic Navigation: A Review of Advanced Endoscopic Vision Technology & \cite{fu2021future}   & IEEE Access & Broad overview of optical Endoscopic and Advanced vision technologies from 1990 to 2020 \\
\hline
12 & A Review on Deep Learning in Minimally Invasive Surgery & \cite{rivas2021review} & IEEE Access& Deep learning applications in MIS like tool detection, segmentation, skill assessment and phase recognition\\
\hline
\end{tabular}
\footnotetext[1]{Book Chapter}
\end{minipage}
\end{table*}

In summary, most of the surveys published in the last five to six years have focused either on broader aspects of surgical navigation tasks, or have restricted their analysis to specific tasks such as instrument detection and tracking or depth estimation; moreover, most of these surveys cover methods based solely on CNNs. However, recently adversarial approaches are getting growing attention in the surgical domain.  
A thorough discussion of relevant surgical navigation data-sets and SOTA for the navigation tasks are still lacking in almost all the previous surveys. Therefore, this survey seeks to bridge this existing gaps. To this end, the major contributions of this review are, 
\begin{itemize}
\item To categorise SOTA surgical navigation tasks for intra-operative guidance in a systematic way
\item To provide an in-depth technical review of some of the most important surgical navigation tasks, specifically, instrument detection, segmentation, tracking and depth perception. 
\item To present an overview of the datasets used for these tasks in terms of data variability, challenging scenarios and task modalities. 
\item To discuss SOTA in terms of real time implementation for intra-operative decision support.
\end{itemize}

\section{Challenges in tool navigation}\label{sec4}
An effective surgical scene understanding framework requires to localize, segment and track the surgical tool throughout the procedure and estimate the depth information to provide contextual information to the surgeon. However, several challenges exist in the development of an image-guided CAI for such surgical navigation tasks. We elaborate upon specific challenges for surgical navigation below. 



\begin{figure*}[h!]
\centering
\captionsetup{justification=justified}
\includegraphics[width=6.4in]{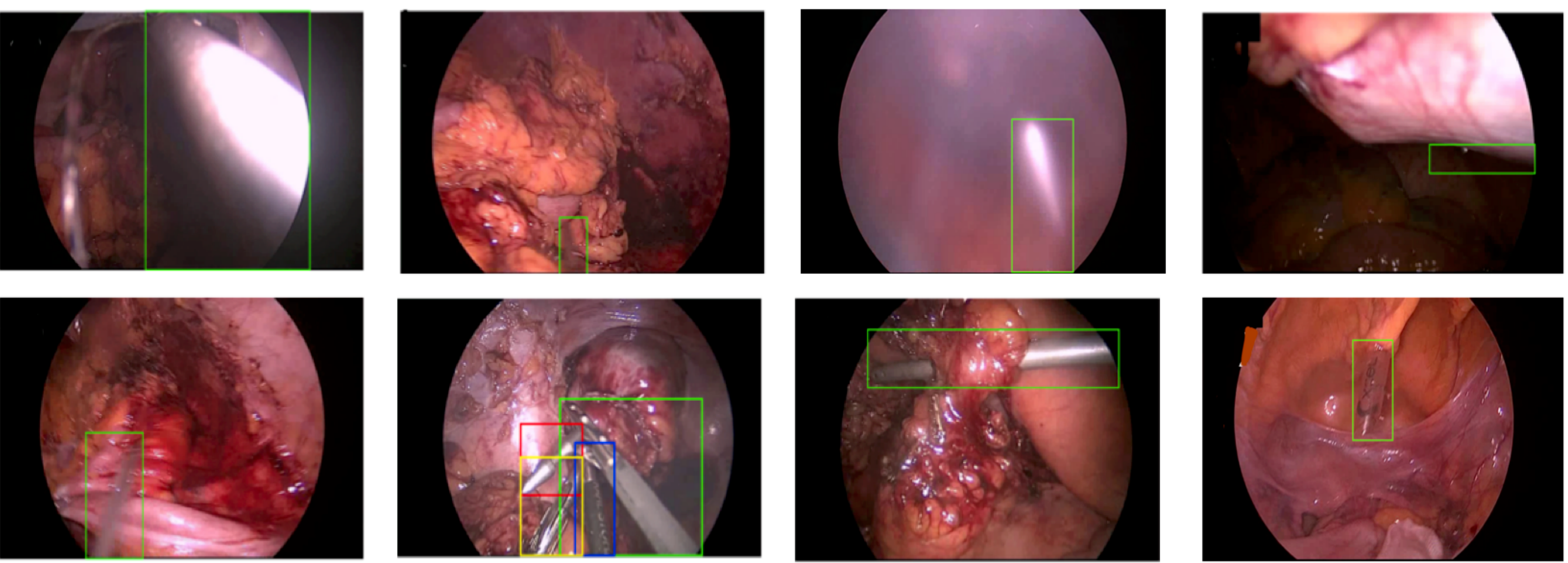}
\caption{Challenging surgical tool images \cite{ceron2022real}. a) partial occlusion due to organ, b) Motion blur, c) occlusion due to smoke, d) Instrument Flare, e) transparent instrument, f) multiple different instruments in the scene, g) underexposed regions with instrument, h) partial occlusions due to blood }
\label{fig3}
\end{figure*}

\noindent\textbf{Surgical tool segmentation}: In contrast to natural images and videos, surgical domain video frames contain several challenging scenarios making it extremely difficult to localize and segment instruments in a robust manner. More specifically, high tissue deformations and occlusions due to the presence of multiple artifacts and blood on the instruments, certain photo-metric artifacts \citep{28} can hamper the performance of the models. Moreover, subtle inter-phase or intra-phase variance, limited field of view of the endoscopic camera \citep{26}, blurriness due to camera motion and gas generated by surgical instruments, specular reflection and scale variation \citep{27} can degrade the segmentation performance. Furthermore, in the case of multiple instruments segmentation, the appearance and shape similarity between different tools, and the presence of tools on the edge of video frames makes it hard to detect and segment the instruments reliably (Fig. \ref{fig3}). Also, the variation in instrument pose may cause changes in the geometry or shape depending on the endoscopic camera field of view.  




Data-specific challenges in DL-based surgical navigation includes a general lack of labeled data for training supervised learning methods, with the addition of class imbalance problems. The class imbalance problem can be present either in foreground-background classes or in the foreground instances . The foreground-background imbalance arises from the fact that fewer image pixels account for the small-sized instruments, while the majority of the image is represented by background pixels.  


The current SOTA tackles the instrument segmentation problem in terms of pixel-wise classification and mostly ignore the global semantic correlations among pixels across subsequent images,  resulting in imprecise feature distribution. 


\noindent\textbf{Surgical tool tracking}: The instrument tracking problem in the AI-based methods have been covered through two-phased approaches in the literature (Detection and Tracking or Segmentation and Tracking). Therefore, the challenges mentioned above relate to instrument localization also impact instrument tracking. Furthermore, motion-blur due to fast moving tools and complex tool trajectories can degrade the tracking performance \citep{du2019patch}. Illumination changes and absence of tools in certain frames due to the removal and reinsertion of the endoscope may also lower the tracking system performance \citep{richa2011visual}. Multi-stage techniques are generally susceptible to high dependence on hyper-parameter tuning. Moreover, window sliding methods for instrument detection can cause missed-detections, since the surgical instruments can come in varying shapes. Also, many Deep Learning based methods such as Fast-RCNN have achieved SOTA performance in tool detection and localization \citep{du2018articulated} but are computationally expensive, introducing inference time penalties.  



\end{sloppypar}


\noindent\textbf{Depth perception}: Depth perception of surgical scenes is a relatively recent topic in the literature. The main idea of these methods is to recover 3D information from the surrounding tissues so as to effectively guide the tool navigation path and be able to reconstruct the internal tissue structure in 3D for better scene understanding and 3D registration between pre- and intra-operative phases. Earlier approaches have used Simultaneous Localization and Mapping (SLAM) and structure from motion (SfM) techniques. As SLAM and SfM approaches are highly dependant on image features struggle to provide accurate results in the surgical domain since endoscopic images contain  texture-less surfaces, and thus less discriminant features. Alternatively, CNN based approaches are have been increasingly used in the recent literature, however, supervised CNN architectures require high amounts of training data, which is time consuming and tedious to obtain. 


\section{Systematic review}\label{sec5}

\subsection{Instrument segmentation}
Intra-operative assistance for the surgeon can be efficiently provided by segmenting the tools in the surgical scene and subsequently classifying the found instances. This task has been achieved in the literature initially by fully supervised methods. Earlier methods used modified FCN and U-Net-based models for instrument segmentation. Typically, the segmentation task can be divided into binary segmentation (separating tools from the background), segmentation of  articulated parts of tool, and a multi-class instrument (instance) segmentation. Since surgical scenes usually contain various artifacts such as specular reflections, few approaches have used attention mechanisms to address the problem. Weakly and semi-supervised techniques have been proposed to address the lack of annotated data, while adversarial approaches have been used to enable the models to generalize from source to target domains. In the next sections, all these methods have been separately reviewed. (Table \ref{tab4} and Table \ref{tab5}) summarise the instrument segmentation SOTA.  

\subsubsection{Supervised architectures}
Semantic segmentation achieved promising results with the introduction of Fully Convolutional Network (FCN) \citep{long:CVPR} architectures. This review starts the discussion of supervised methods by introducing FCN-based methods  used in surgical tool segmentation. Moreover, efforts towards developing a model to produce real-time inference has also been reviewed. Recently, meta-learning is being used for instrument segmentation which is also discussed at the end of this section. \\

\begin{figure*}[t!]
\centering
\includegraphics[width=6.2in]{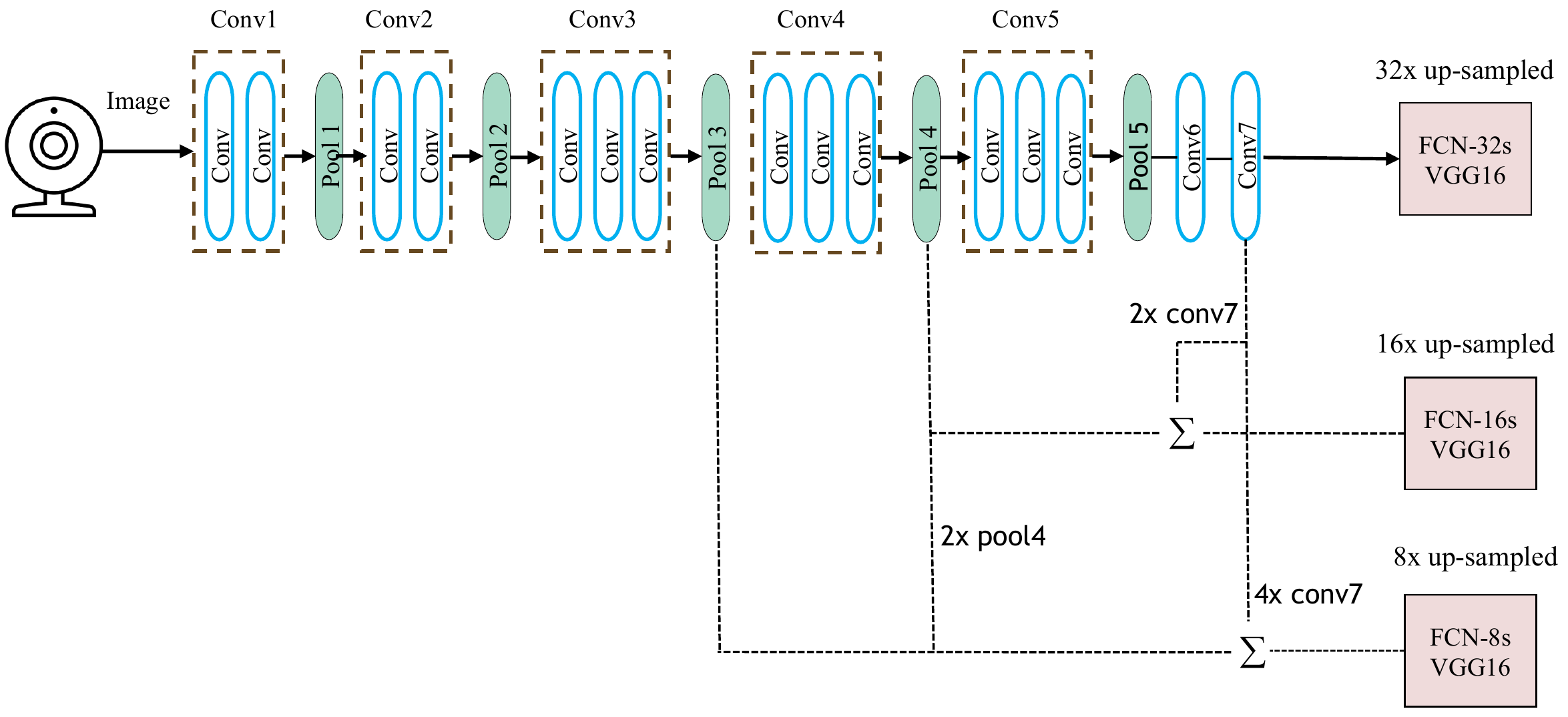}
\caption{Different architectures of FCN}
\label{fig4}
\end{figure*}

\noindent\textbf{Methods based on FCNs}: DL-based methods like FCNs have outperformed the traditional hand-crafted feature models like SIFT \citep{SIFT:CV}, SURF \citep{SURF:CV} and HoG. The FCN architecture works by replacing the fully connected layer of the classification model to use a convolutional layer to get a pixel-wise classification. Additionally, the FCN uses skip connections to fuse semantic information from coarse layers with localization information from fine layers. In this way, FCNs offered a new way to approach semantic segmentation problem (different FCN variants are shown in Fig.\ref{fig4}). However, FCNs suffer from several issues. First, the  generated output segmentation mask is eight times smaller than the input, causing a loss of features and degradation of performance on the edges. Second, the loss function used in the original work does not work  well on unbalanced datasets. Finally, FCN lacks real-time capabilities, which is essential in applications such as surgical navigation. 

Several preceding works \citep{U-Net:MICCAI} are based on FCN modifications to improve segmentation performance in challenging scenarios and address the lack of annotated data by redesigning either the encoder or decoder network. Based on FCN, U-Net \citep{U-Net:MICCAI} has been the most popular among the medical image segmentation SOTA. It has been designed to work with less training data samples. The encoder convolutional blocks of U-Net downsample the input by a factor of two while similar blocks in the decoder upsample them to get an output of the same resolution as the input.

\begin{figure*}[t!]
\captionsetup{justification=justified}
\centering
\includegraphics[width=\textwidth,height=3in, keepaspectratio]{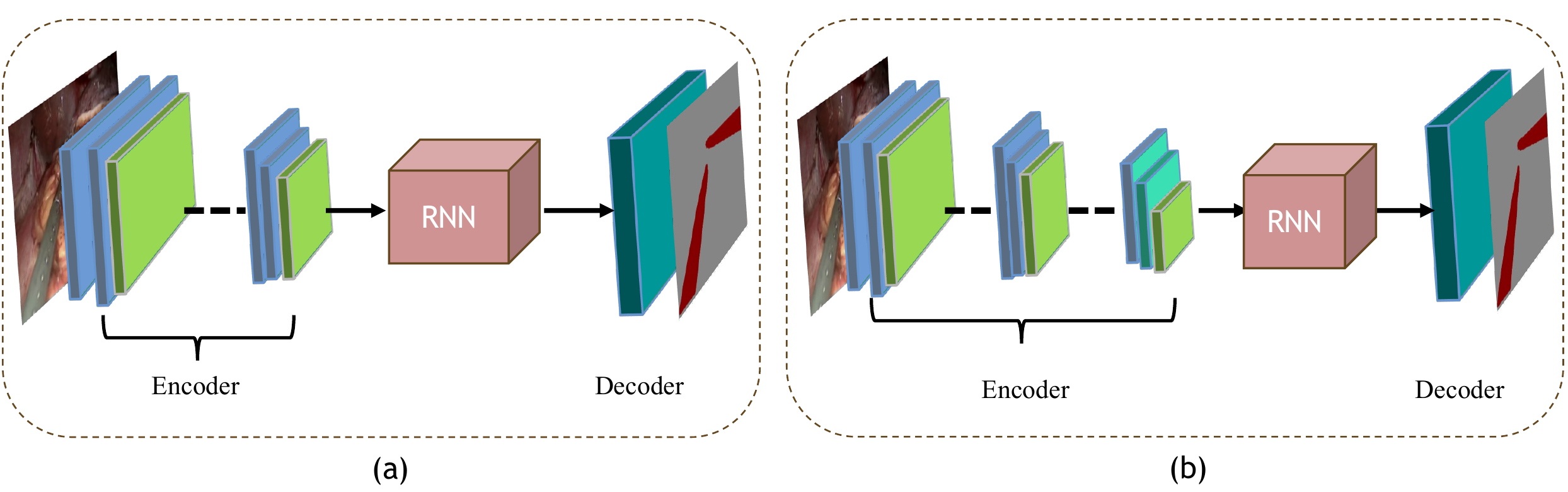}
\includegraphics[width=\textwidth,height=3in, keepaspectratio]{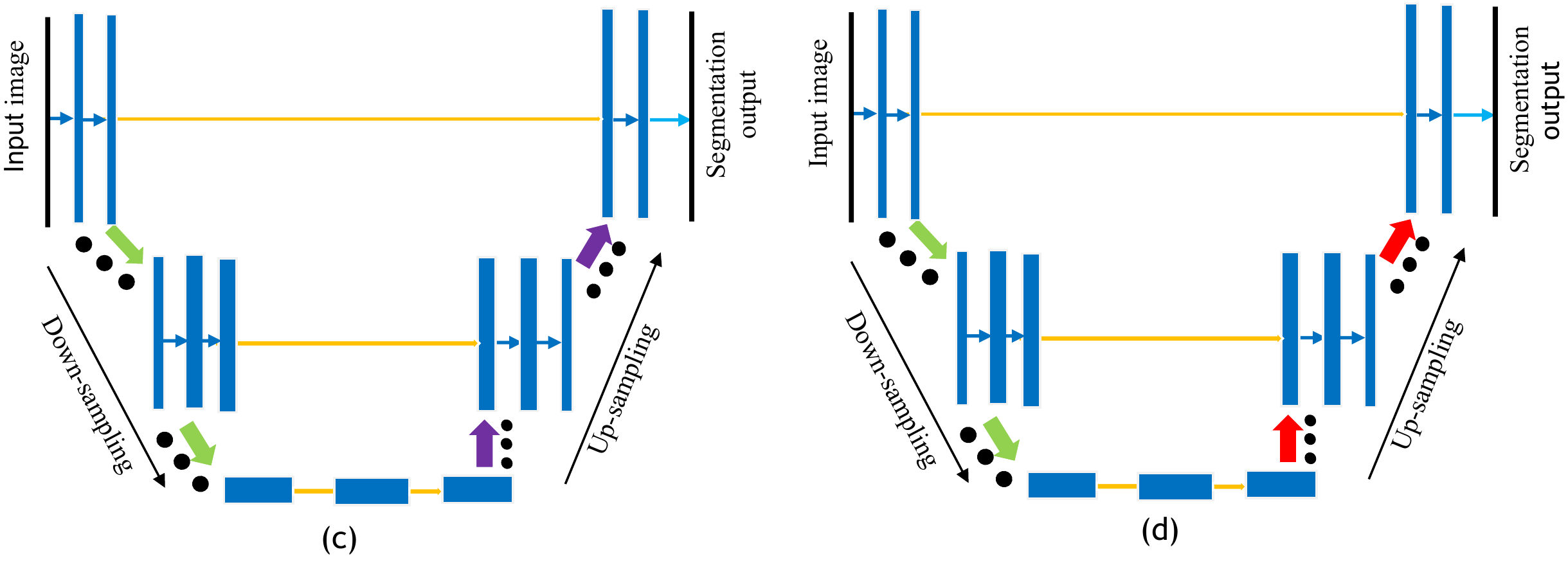}
\includegraphics[width=\textwidth,height=3in, keepaspectratio]{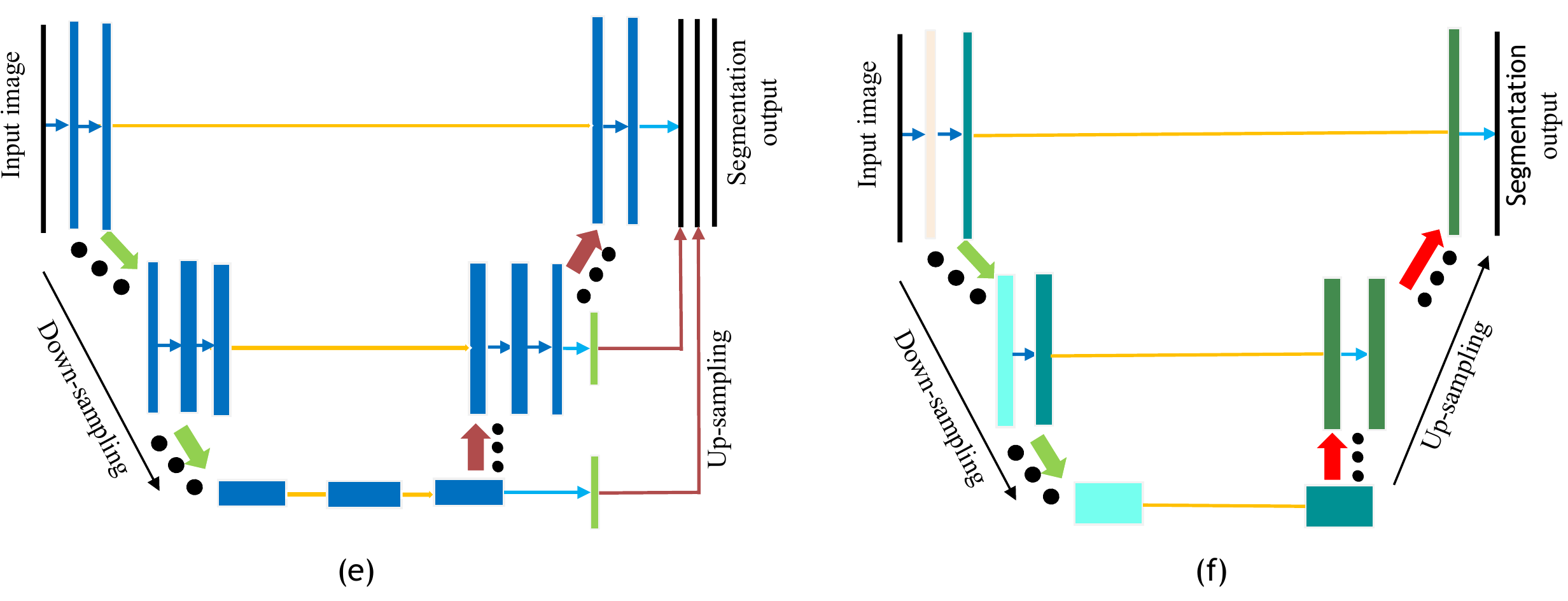}
\includegraphics[width=\textwidth,height=3in, keepaspectratio]{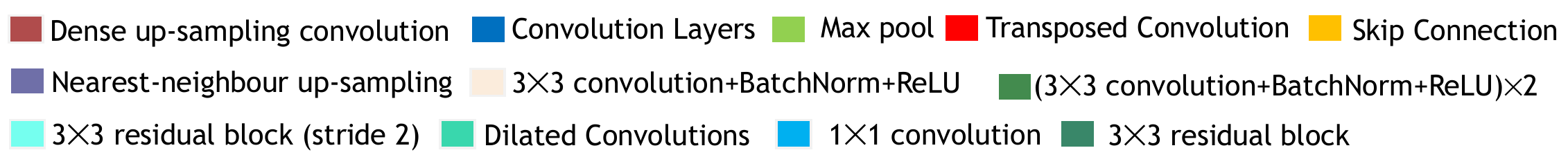}
\caption{Various modifications of FCN and U-Net used in the literature. a) RNN added to FCN, b) Addition of Dilatd convolutions into FCN, c) Classicial U-Net architecture, d) U-Netplus:Nearest neighbour up-sampling instead of transposed convolution ,e) Use of Dense up-sampling convolution, f)OR-U-Net: Use of residual networks} 
\label{fig5}
\end{figure*}

Several U-Net modifications have been proposed in the literature (Fig.\ref{fig5}), such as the addition of residual connections \citep{apostolopoulos:MICCAI} or the use of dense connections \citep{gibson:TMI}. Other modifications have been proposed as well. For instance, Chen \emph{et al.} added sub-pixel layers in the U-Net model to improve its performance on low light settings \citep{chen:2018}. \cite{shvets:ICMLA}, the winning team of the EndoVis 2017 Instrument segmentation sub-challenge and the first to propose multi-class segmentation, proposed a modified version of the classical U-Net architecture for instrument segmentation in both binary and multi-class setting, by using pre-trained encoders. The model uses a deconvolution operation, which works just opposite to that of convolution. For instance, instead of mapping from 3×3 to 1, 1 to 3×3 mapping is performed to get an upsampled output. But the problem with this approach is that it requires additional parameters and weights making it slow for an end-to-end training framework and for low overhead inference. Furthermore, deconvolution blocks can cause 'uneven overlap' problems and introduce undesirable artifacts (in a similar way to the checkerboard kind of patterns in different colors and shapes as pointed out in \cite{radford:arxiv} and \cite{salimans:ANIPS}. More recently, U-NetPlus \citep{hasan:EMBC}, a modification of U-Net tries to minimize the occurrence of those artifacts. This model uses VGG as an encoder with batch-normalized  pre-trained weights, replacing the transposed convolution layers with nearest-neighbour interpolation. The pre-trained encoder helps speed up the convergence and the decoder with interpolation removes artifacts. The results show a slight improvement in binary segmentation but still instrument parts and type segmentation needs further refinement. 
Inspired by the work of \citep{xie:CVPR}, authors of \cite{yu:DI} proposed holistically nested U-Net architecture in which they replaced the transposed convolution network of the upsampling layers in the classical U-Net with dense up-sampling convolution DUC which was initially used in \cite{wang:WACV}. The DUC helps to get an output segmentation image of the same dimensions as the input. The major limitation of this model is that the output segmentation masks have coarse boundaries. 

ToolNet \citep{toolnet:IROS} followed the FCN architecture by addressing some of its limitations like upsampling issues and lack of real-time performance. This work is based on an FCN-8s enhancement by fusing multi-scale feature fusion in a cascaded arrangement to further refine edges of surgical instrument segmentation output. It also uses a simplified decoder to make the network lightweight and able to run in real time (29 FPS). The number of model  parameters are 15 times less as compared to the FCN-8s. While the model produces some useful results, it suffers from limited feature receptive fields which degrades model performance in cases of complex surgical scenes. Most of the architectures used in the literature tend to be shallow: i.e., they do  not extract more discriminatory features and suffer from small receptive fields. Trying to address these challenges, authors in \cite{kadkhodamohammadi:Springer} redesigned the encoder-decoder architecture for surgical scene segmentation. The encoder part is implemented using the Xception Model \citep{xception:CVPR} which incorporates deeper convolutions like those used in DeepLabv3+ \citep{DeepLabv3+:ECCV} to extract rich feature representations. On the other hand, the decoder part is designed to fuse multi-scale features and adapted to combine more meaningful features. The rationale is to avoid a deeper decoder network and instead re-use features extracted by a deeper encoder. The model yielded promising results on the endoVis15 dataset. Instead of a single encoder, StreoScenNet \citep{streoscennet:MI} uses an ensemble of multiple pre-trained encoders and a decoder with new sum-skip connections to better perform instrument segmentation in all three modalities. The model uses both right and left images of a stereo endoscopic camera. The proposed model achieved 1.6 points improvement in instrument type segmentation, whereas in other tasks did not perform well. 

Another technique to boost the performance of FCN is to use combinations or more complex blocks, like the one used in \cite{garcia:IWCAR}. The proposed model combines optical flow tracking and a recurrent neural network (RNN) to implement both real-time and non real-time binary instrument segmentation. Non real-time based on only deep learning gives 89.6 outperforming SOTA by 3.2 points in  balanced accuracy while real-time adds optical flow to the model to produce average balanced accuracy of 78.2. The model is based on fine-tuning an FCN-8s. Since the FCN model is not suitable for real-time applications, optical tracking, and affine transformation is employed to ease the burden of performing feature extraction in every frame thus reducing the computational complexity and making the network perform in real-time. Most of the methods based on FCNs use simple concatenation \citep{U-Net:MICCAI,long:CVPR,laina:MICCAI} to  perform multi-scale feature fusion which is a very simple approach. Instead of that, LSTM \citep{LSTM:NC} in the decoder block along with deeper architecture in encoder network is proposed in \cite{cfcm:MICCAI} to smartly add features from different scales. The proposed model is based on the idea that LSTM having a better memory mechanism can handle and retain important features extracted at various stages in the encoder network in memory. The network achieves balanced accuracy of 97.8\% in binary segmentation and shows similar performance in instrument parts segmentation. The major strength of the framework lies in its robustness to specular noise and accurate grasper segmentation.

Generally, the predictions of the FCN based models reduce the image dimensions significantly in the encoder part (for instance VGG or ResNets reduce feature map size by 32). Segmentation models must maintain high-resolution feature maps in the model for dense prediction, but this has the undesirable effect of increasing the training time. Therefore, there exists a trade-off between training time and high resolution. The Atrous convolution block was added to U-Net by the authors in \cite{gu:TMI} inspired by inception ResNet frameworks \citep{szegedy:AAAI} to make the CNN denser and wider. In an another  work \citep{Evan:PAMI}, it was shown that using an up-sampling deconvolutional network with a factor of 32, the lost dimensions can be recovered and original image dimensions can be restored but at the cost of very coarse segmentation boundaries which are often not desirable. Several works have tried to address this inherent problem of FCNs. One method \citep{Evan:PAMI} adopted is to use feature fusion between corresponding encoder and decoder layers to better predict segmentation masks and compensate the coarse boundary problem. Another technique \citep{yu:arXiv} is by modifying the encoder downsampling network (like VGG16) by removing some of the pooling layers to avoid dimension reduction in feature maps or making the strides to one instead of two in some convolutional layers of ResNet to stop further downsampling. In this regard, authors in \cite{deeplab:PAMI} employ dilated convolutions to tackle this problem.  Another approach in this direction, apart from solving the limited receptive field problem of FCN and the over-segmentation problem of FCNs, a hybrid arrangement of CNN-RNN is proposed in \cite{attia:ICSMC}. Over-segmentation occurs due to the lack of dependency between patches of images implemented in convolutional layers. To overcome the problem, recurrent layers are added into the proposed pipeline. Here, the convolutional layers model the deep semantic features while the recurrent layers extract the spatial local and global dependencies. In order to make the model computationally efficient, strided convolutions are utilized in the reconstruction part. The model tested on EndoVis 2015 dataset obtained 93.3\% balanced accuracy in binary instrument segmentation. 

Combining deep residual CNN with FCN along with dilated (atrous) convolutions to account for the more-than-required reduction in feature map downsampling, authors in \cite{pakhomov:IWML}, extend for the first time, the robotic instrument segmentation pipeline to multi-instrument segmentation. In the proposed framework, atrous convolutions are seen as a useful alternative to the deconvolutional network and Skip architecture of FCN-8s. This approach results in 4\% improvement balanced accuracy as compared to SOTA. \\

\noindent\textbf{Using temporal information}:
Several methods proposed in the literature consider a surgical video frame as a static image not utilizing temporal cues in segmentation or tracking of instruments. Methods that Model the motion between frames have been approached using optical flow in the literature \citep{garcia:IWCAR,jin:MICCAI} which has a high computational cost. To address this problem and develop a real-time instrument segmentation pipeline, dual memory network DMNet has been proposed that integrates spatio-temporal local and global information. The results on two public datasets indicate the effectiveness of the approach as it gets a 61.03\% mean dice score with an FPS of 38 as compared to the fastest model LWANet \citep{lwanet:ICRA} as it gets 49.79\% which has 76 FPS.  Computer vision studies on natural images have validated the idea of integrating a long range global context in performance improvement like in action recognition \citep{wu:CVF}, super-resolution \citep{yi:CVF} and video object segmentation \citep{feelvos:CVF}. Other studies incorporating temporal cues are focused on surgical workflow recognition and tool presence detection \citep{martel:MICCAI,van:ICRA,martel2021:MICCAI,jin2017:TMI}. The instruments position and shape are determined through optical flow in the current frame with the help of temporal cues by propagating predictions from the previous frame in an unsupervised way in the work proposed by \cite{jin:MICCAI}. In another study \citep{mishra:CVPR}, the authors combine LSTMs \citep{LSTM:NC} with CNNs to capture temporal context for multi-class instrument segmentation. 

\noindent\textbf{Real-time architectures}:
It is essential for a framework to work in real-time to provide valuable insights about the tool's presence and its trajectory during the procedure. Some studies \citep{lwanet:ICRA} focus on the real-time aspect of the instrument segmentation domain which has largely been under-studied or the accuracy being compromised because of the inference time. To this end, \cite{pakhomov2020:MICCAI} proposed binary and instrument parts segmentation model that uses lightweight residual network. This approach is essentially an improvement over the work in \cite{pakhomov:IWML} that used dilated residual connection which is very computationally expensive due to a large number of filters in the last layers. The proposed model is pre-trained on ImageNet dataset and exhibits low latency and occupies small GPU memory along with a mechanism to search optimal dilation rates. The model yields competitive segmentation performance with a major inference speed of 125 FPS. 
Various lightweight architectures like ShuffleNet \citep{shufflenet:CVF} and MobileNet \citep{MobileNet:CVF} have been proposed in the literature to meet the requirements of real-time implementation and reduce the computational burden but only few studies have incorporated them into surgical navigation tasks where the major focus has been on improving the segmentation performance only. A feature fusion-based approach has been proposed by \cite{islam:IEEE} for the instrument segmentation outperforming state-of-the-art models like ICNet \citep{icnet:ECCV}, PSPNet \citep{pspnet:CVPR} and LinkNet \citep{linknet:VCIP}. In another work \citep{qin:IEEE}, the limited rotation invariant performance of deep neural networks are addressed by using a multi-angle feature fusion approach without increasing the number parameters. This study focuses on sinus surgical instruments. In a two-stage framework \citep{sun:IEEE}, Lightweight MobileNetV3 \citep{MobileNet:CVF} is combined with ghost modules \citep{ghostnet:CVF} and segmentation head \citep{Howard:ICCV} to develop improved real-time instrument segmentation with inference speed of 37 FPS. ToolNet \citep{toolnet:IROS} proposes a simplified decoder network to reduce the computation time and fasten the inference. Authors in \cite{islam:MICCAI} propose multi-task learning framework with light decoder based on saliency features to suppress unwanted regions and highlight salient features. A trade-off between accuracy and performance is can be observed in the proposed lightweight multi-feature fusion architecture \citep{islam:IEEE} for instrument segmentation.  

\noindent\textbf{Addressing lack of data}: One of the most common problem in segmentation tasks is the lack of annotated data especially when it comes to medical data. Typical approaches to overcome this problem have been based on the use ImageNet pre-trained encoders, which has been applied in robotic instrument segmentation \cite{EndoVis17:arXiv}. Some recent studies like \citep{he:CVF} suggest that the benefits of using pre-trained networks may be far less than expected, particularly for larger target domain datasets. In that connection, residual blocks \citep{he:ECCV} were added to the encoder network of U-Net to train the segmentation model without using pre-trained encoder on the assumption that EndoVis 17 dataset \citep{EndoVis17:arXiv} is sufficient to train the model without overfitting \citep{isensee2021nnu}. However, no evidence or experiment has been performed to validate the claim. Another approach to compensate for the lack of data was proposed by \cite{colleoni2020synthetic:MICCAI} where authors use multi-modal data and train an FCN model on both simulated and real surgical images for improved performance. In this work, data generated from robot simulator and real laparoscopic camera were used to create a custom dataset with segmentation labels and kinematic information. Results from the model show encouraging output as experiments were performed by adding blood and noise on simulated tool videos. 

\noindent\textbf{Instrument segmentation as instances}:
In contrast to semantic segmentation, instance segmentation  has some inherent advantages. Semantic segmentation does not answer questions like how many instances of a particular class are present in the image or whether some of them are occluded. Instances can help in further applications like pose estimation \citep{allan:TMI} and tracking \citep{kurmann:MICCAI}. Preserving the global visual features using instance based approaches aid in misclassification problems for multi-class tool segmentation. Authors in \cite{kurmann:IJCARS} propose a novel surgical tool instance segmentation approach by producing three different pixel-wise representations of the input image-segmentation mask, offset regression and centroid heatmap. The proposed framework does not use bounding boxes since they can cause several problems like overlapping of one bounding box over another or issues on the boundary of the image. The model projects an improvement in classifying instrument types and parts in compassion to semantic segmentation. Another work \citep{kletz:CBMI} has focused on binary and multi-class instrument segmentation as instances in laparoscopic gynecology. These approaches generally add an extra step to separate different objects from one another. The work comprises 11 instruments, in multi-class instance segmentation setting, where instruments such as needles, bipolar have shown to be the hardest to segment owing to their complex geometry and thin structure respectively. Multi-class instance segmentation was performed by adding temporal information module to MaskRCNN in \cite{isinet:MICCAI}. Temporal information helped preserve the instruments' identity across frames while the model showed improved generalization when trained on both EndoVis 2017 and 2018 datasets. Another instance-based instrument parts and type detection and segmentation approach is based on MaskRCNN refinement by setting some rules to generate regions of interests ROIs \citep{kong:IJCARS}. In this framework, ROIs are generated to indicate a positive example regarding a frame containing tool if the anchor contains an intersection over union (IoU) overlap of greater than 0.5 and IoU of the frame is between 0.3 and 0.5. This is done so as the model does not miss small appearing instrument parts like jaws.


 
\noindent\textbf{Methods using attention mechanism}:
Proper lighting conditions are required while performing MIS procedure which undesirably lead to specular reflections, appearance of shadows due to angle of illumination. The lighting variation changes the visual features of the instrument making it difficult to segment. Being in continuous movement, in some frames, part of the instrument may appear or number of background pixels are usually much higher than foreground ones hindering the segmentation models to learn discriminatory features. variation in instrument can cause apparent geometry change of the instrument in certain frames. In that scenario, attention guided networks \citep{li:arXiv,fu:CVF} have been proposed mimicking the human attention system and try to take into account neighbouring pixels and global context to better segment the tool. A dual attention network \citep{fu:CVF} modeled semantic dependencies between channels and position using channel and position attention networks. In this work, Position and Channel attention blocks were fused to extract information simultaneously. Semantic dependencies were modelled by squeezing the global context features into a vector in squeeze-excitation block in the non-local block. A similar approach was used in the Progressive attention guided module \citep{zhang:CVPR}. Attention networks have been efficiently integrated with FCNs in the literature to perform semantic segmentation. For instance, replacing simple skip connections in U-Net with attention modules, the Refined attention segmentation network RASNet \citep{ni:EMBC} was proposed to make use of deep semantic features from the encoder network efficiently. Authors validated the effectiveness of attention fusion module AFM by showing 5 points improvement in dice score over the network without using AFM. In another encoder-decoder network \citep{lwanet:ICRA}, a lightweight decoder is proposed for semantic instrument segmentation. It contains depth-wise separable and transposed convolutions and an attention fusion block. The model exhibited a competitive performance in terms of segmentation and computational cost, while using comparatively lower number of trainable parameters. In another attempt to highlight target regions, PAANet \citep{ni:AAAI} employs double attentive module DAM for the position and channel dependencies and Pyramid Upsampling Module PUM for fusing multi-scale attentive features. The proposed architecture produces significant performance improvement on Cata7 and EndoVis 2017 challenge dataset. An attention module was added to the decoder in the multi-task framework in \cite{islam:MICCAI}  to perform instrument type segmentation and tracking.\\ However, the model does not incorporate temporal cues, which are essential for a long range activity. In a further refinement of the multi-task network, ST-MTL \citep{islam2021:MIA} proposes a novel decoder design by task-aware spatio-temporal unit along with saliency map generation. The proposed model addresses the convergence problem of multi-task networks. Results were very encouraging but the model uses a large number of parameters which impacts computational performance. In contrast to open surgery, cataract surgery requires additional lighting which causes strong specular reflection problem. RAUNet \citep{ni2019raunet} addresses this problem along with class imbalance issue of small-size cataract surgery instruments. 
 
Surgical instrument segmentation suffers from scale variation issues, since it is affected by the variations in tool shapes and sizes. Several approaches in the literature address this problem by aggregating multi-scale features  of the target objects. These methods focus on enlarging the receptive field to get better contextual information. In that regards, PSPNet \citep{pspnet:CVPR} proposes spatial pyramid pooling for feature aggregation, dilated convolutions having varied dilation rates were employed in \citep{denseaspp:CVPR,deeplab:PAMI} to generate features with varying receptive field in atrous spatial pyramid pooling ASPP architecture. Another work in \cite{chen2017:arXiv} blends dense connections in ASPP to mitigate scale variations in target objects. PUM \citep{ni:AAAI} addresses the scale variation problem. A pertinent point to mention here is that majority of studies have ignored a subtle point. A pixel may not always require a larger receptive field. For instance, larger receptive fields can be beneficial for a pixel at the center but counter-productive at the boundary. Moreover, attention based techniques use flexible-length convolutional kernels to obtain long-range context which increases the computational time and  is not required since only the information from neighbouring pixels is useful. Additionally, in that pipeline, models output comes as one-eighth of the input  size which is then up-sampled using some interpolation technique that is not an efficient way and may produce misaligned outputs. Inspired by the architecture in \citep{denseaspp:CVPR}, CycleASPP \citep{qu:CVF} proposed both forward and backward connections between atrous convolution layers to better adjust the receptive fields. BARNet \citep{ijcai2020p116} addresses the problem by using an adaptive receptive field module. With a focus on both real-time performance and improved accuracy, authors in \cite{ceron2022real} proposed lightweight architecture with convolutional block attention network to perform surgical instrument segmentation. 

\noindent\textbf{Meta-learning in instrument segmentation}:

 Machine learning and deep models have been greatly successful in several applications \citep{he2016deep:CVPR, silver2016:nature, devlin2019kt}. Still, there are clear limitations \citep{marcus2018:arXiv}, core among them is the reliance on a vast amount of labelled data for training a model to be robust and generalizable. This limits their usability in several areas where collections, annotation of data is cumbersome or tedious or heavy computing resources are not available \citep{altae2017:ACS}. In this scenario, meta-learning offers a viable alternative where a model learns over multiple episodes and uses that experience to perform better future predictions. 

Computer vision has been a major beneficiary of meta-learning paradigm because of its usability in few-shot learning (FSL). FSL has been quite challenging specifically for large ML models where vast data availability is the deciding factor for performance. In other cases, training those models with few data as in the surgical domain, leads to model overfitting problems or non-convergence. Meta-learning has been extremely successful in training DL models with small amounts of data in many computer vision applications. Meta-learning in few shot object segmentation has been quite helpful in the form of hyper-network based meta-learners since obtaining pixel-wise labeled data is quite challenging \citep{shaban2017:arXiv}. Meta learning has also been used in simulation to real domain adaptation where the inner-level network learns the simulation domain features while the outer-level computes the model performance on real-world data. 
Meta-learning in the surgical domain has been explored by \cite{zhao2021anchor:MIA} where they investigate meta-learning for domain adaptation into two scenarios. 1) A large domain shift (general to surgical data) 2) Small domain shift (public to in-house data). A refined two-stage meta-learning approach named anchor-guided online meta adaptation (AOMA) is proposed in which model initialization is learned in the first stage from the easily accessible source data while second stage implements fast adaptation requiring only first frame annotation from target data.The model achieved a significant enhancement in adaptation run-time performance with only 1.8 seconds to adapt to new surgical videos. In another similar setup \citep{zhao2021one}, MDAL, a meta-learning based dynamic online adaptive learning scheme proposes an adaptive instrument segmentation framework which can adapt from one source domain to many target domains requiring single first frame from the target domain. MDAL produces 75.5\% IoU and 84.9\% Dice on new domains. 

\begin{sidewaystable*} [!htbp]
\tabcolsep7.5pt
\footnotesize
\begin{minipage}{\textheight}

\caption{Review of Instrument segmentation SOTA}\label{tab3}
\begin{center}
\begin{threeparttable}
\begin{tabular}{@{}lllllllll@{}}\toprule
\multicolumn {5}{c}{\textbf{Training}} &&&\multicolumn{1}{c}{\textbf{Test}} \\\cmidrule(l){3-7}\cmidrule(l){8-8}

\textbf{Ref.}&    \textbf{Year}& \textbf{Architecture} & \textbf{Tool} & \textbf{Dataset} & \textbf{Data} & \textbf{Technique} & \textbf{Data} & \textbf{Application Task}\\
\hline
\cite{garcia:IWCAR}&2016&FCN-8s+&Robotic&EndoVis15,NST$^{\rm a}$&Real&Supervised&Real& Segmentation(B$^{\rm c}$)\\
&&Optical Flow&&FFT$^{\rm b}$&&&&\\
\cite{pakhomov:IWML}&2017&ResNet+atrous&Robotic&EndoVis15&Real&Supervised&Real&Segmentation(B,I$^{\rm c}$)\\
\cite{attia:ICSMC}&2017&CNN+RNN&Robotic&EndoVis15&Real&Supervised&Real&Segmentation(B$^{\rm c}$)\\
\cite{toolnet:IROS}&2017&ToolNet&Robotic&DVR&Real&Supervised&Real&Segmentation(B$^{\rm c}$)\\
\cite{mishra:CVPR}&2017&CNN+LSTM&Rigid&m2cai16-tool&Real&Supervised&Real&Segmentation(I$^{\rm c}$)\\
\cite{shvets:ICMLA}&2018&Ternaus11, ,&Robotic&EndoVis-17&Real&Supervised&Real&Segmentation(B,P,I$^{\rm c}$)\\
&&Ternaus16,&&&&&&\\
&&LinkNet34&&&&&&\\
\cite{cfcm:MICCAI}&2018&ResNet +&Rigid&EndoVis15&Real&Supervised&Real&Segmentation(B,P$^{\rm c}$)\\
&&Conv LSTM&&&&&&\\
\cite{ross:IJCAS}&2018&ResNet,U-Net$^{\rm d}$*&Robotic&EndoVis17&Real&Self-S\tnote{5}&Real&Segmentation(B$^{\rm c}$)\\
\cite{hasan:EMBC}&2019&U-NetPlus&Robotic&EndoVis-17&Real&Supervised&Real&Segmentation(P,P,I$^{\rm c}$)\\
\cite{kletz:CBMI}&2019&ResNet101+  &Rigid&Custom&Real&Supervised&Real&Segmentation(B,P$^{\rm c}$)\\\
&&MaskRCNN&&&&&&\\
\cite{kadkhodamohammadi:Springer}&2019&Xception +FAD\tnote{1}&Rigid&EndoVis15,LSG$^{\rm e}$ &Real&Supervised&Real&Segmentation(P,I$^{\rm c}$)\\
\cite{streoscennet:MI}&2019&StreoScenNet&Robotic&EndoVis17&Real&Supervised&Real&Segmentation(B,P,I$^{\rm c}$)\\
\cite{jin:MICCAI}&2019&MF-TAPNet&Robotic&EndoVis17&Real&Supervised/&Real&Segmentation(B,P,I$^{\rm c}$)\\
&&&&&&SS$^{\rm d}$&&\\
\cite{ni:EMBC}&2019&RASNet&Robotic&EndoVis17&Real&Supervised&Real&Segmentation(I$^{\rm c}$)\\
\cite{islam:MICCAI}&2019&ResNet&Robotic&EndoVis17&Real&Supervised&Real&Segmentation(B,I$^{\rm c}$), \\
&&&&&&&&Tracking\\
\cite{fuentes:IJCARS}&2019&DeepLabv3+&Robotic,&ENdoVis15, &Real&WS$^{\rm f}$&Real&Segmentation(B,P,I$^{\rm c}$)\\
&&&Rigid&LSG$^{\rm e}$, GB $^{\rm g}$&&&&\\

\cite{lee:IET}&2019&DCNN&Rigid&Private&Phantom&WS$^{\rm f}$&Real&Segmentation(B$^{\rm c}$)\\
&&&&&&&&Tracking\\
\cite{islam:IEEE}&2019&CNN+Residual&Robotic&EndoVis17&Real&Auxilary/ &Real&Segmentation(B,P,I$^{\rm c}$)\\
&&&&&&Adversarial&&\\
\cite{yu:DI}&2020&Modified U-Net&Robotic&EndoVis-17&Real&Supervised&Real&Segmentation(B$^{\rm c}$)\\
\cite{isensee:arXiv}&2020&OR-U-Net&Robotic&EndoVis17&Real&Supervised&Real&Segmentation(B$^{\rm c}$)\\
\cite{isinet:MICCAI}&2020&ISINet&Robotic&Endovis17, 18&Real&Supervised&Real&Segmentation(I$^{\rm c}$)\\
\cite{qin:IEEE}&2020&MAFA+Deeplabv3 & Rigid, &Sinus Surgery C,L &Real&Supervised&Real&Segmentation(B$^{\rm c}$)\\
&&+TernausNet16&Robotic&EndoVis17&&&&\\
\cite{pakhomov2020:MICCAI}&2020&Light ResNet18&Robotic&EndoVis17&Real&Supervised&Real&Segmentation(B,P$^{\rm c}$)\\
\cite{ijcai2020p116}&2020&BARNet&Rigid, &Cata7, EndoVis17&Real&Supervised&Real&Segmentation(I$^{\rm c}$)\\
&&&Robotic&&&&&\\
\bottomrule
\end{tabular}
\begin{tablenotes}
$^{\rm a}$ Neuro Surgical Tools; $^{\rm b}$ FetalFlex Tool ; $^{\rm c}$ B=Binary, P=Parts, I=Instance ;$^{\rm d}$ Modified U-Net; $^{\rm e}$ Laparoscopic Sleeve gastrectomy; $^{\rm f}$ Weakly Supervised ; $^{\rm g}$  Gastric bypass dataset with “stripes”;
\end{tablenotes}
\end{threeparttable}
\end{center}
\end{minipage}
\end{sidewaystable*}

\begin{sidewaystable*}[!htbp]
\footnotesize
\begin{minipage}{\textheight}

\caption{Review of Instrument segmentation SOTA}\label{tab4}
\begin{center}
\begin{threeparttable}
\begin{tabular}{@{}lllllllll@{}}\toprule
\multicolumn {5}{c}{\textbf{Training}} &&&\multicolumn{1}{c}{\textbf{Test}} \\\cmidrule(l){3-7}\cmidrule(l){8-8}

\textbf{Ref.}&    \textbf{Year}& \textbf{Architecture} & \textbf{Tool} & \textbf{Dataset} & \textbf{Data} & \textbf{Technique} & \textbf{Data} & \textbf{Application Task}\\
\hline
\cite{liu2020:MICCAI}&2020&CNN+U-Net&Robotic&EndoVis17&Real&Unsupervised&Real&Segmentation(B$^{\rm a}$)\\
\cite{sahu2020:MICCAI}&2020&DNN+&Rigid,&Sim$^{\rm b}$, SimCholec80,&Real,&UDA$^{\rm c}$&Real&Segmentation(B$^{\rm a}$)\\
&&Ternaus11&Robotic&EndoVis15&Simulated&&&\\
\cite{zhao:MICCAI}&2020&FlowNet2.0,&Robotic&EndoVis17&Real&SS$^{\rm d}$&Real&Segmentation(B,P,I$^{\rm a}$)\\
&&ConvLSTM&&&&&\\
\cite{lin2020:IROS}&2020&LC-GAN&Rigid&Sinus Surgery&Real, &Adversarial&Real,&Segmentation(B$^{\rm a}$)\\
&&&&&Synthetic&&Synthetic&\\
\cite{colleoni2020synthetic:MICCAI}&2020&FCNN&Robotic,&Custom&Real,&Supervised&Real,&Segmentation(B$^{\rm a}$)\\
&&& Rigid&&Simulated&&Simulated&\\
\cite{ni:AAAI}&2020&PAANet&Rigid, &Cata7, EndoVis17&Real&Supervised&Real&Segmentation(I$^{\rm a}$)\\
&&&Robotic&&&&&\\
\cite{lwanet:ICRA}&2020&LWANet&Rigid,&Cata7, EndoVis17&Real&Supervised&Real&Segmentation(I$^{\rm a}$)\\
&&&Robotic&&&&&\\
\cite{kurmann:IJCARS}&2021&CNN&Robotic&EndoVIs17&Real&Supervised&Real&Segmentation(P,I$^{\rm a}$)\\ 
\cite{kurmann:IJCARS}&2021&CNN&Robotic&EndoVIs17&Real&Supervised&Real&Segmentation(P,I$^{\rm a}$)\\ 
\cite{wang:MICCAI}&2021&DWANet&Robotic&EndoVis 17,18&Real&Supervised&Real&Segmentation(P,I$^{\rm a}$)\\
\cite{sun:IEEE}&2021& GhostNet +&Robotic&EndoVis17&Real&Supervised&Real&Segmentation(B,P$^{\rm a}$)\\
&&MobileNetv3&&&&&&\\
\cite{kong:IJCARS}&2021&MaskRCNN$^{\rm e}$&Robotic&EndoVis17, in-house&Real&Supervised&Real&Segmentation(P,I$^{\rm a}$)\\
\cite{sanchez:MICCAI}&2021&EfficientDet&Robotic&EndoVis18&Real&WS/SS$^{\rm d}$&Real&Segmentation(I$^{\rm a}$]\\
\cite{su2021:sensors} &2021&CycleGAN$^{\rm e}$&Rigid&Sinus Surgery&Real,&Adversarial&Real,&Segmentation(B$^{\rm a}$)\\
&&&&&Synthetic&&Synthetic&\\
\cite{colleoni:IRAL}&2021&CycleGAN$^{\rm e}$, MUNIT$^{\rm e}$,&Robotic&EndoVis15,17,20&Real,&Adversarial&Real,&Segmentation(B$^{\rm a}$)\\
&&U-Net&&&Synthetic&&Synthetic&\\
\cite{zhang2021:IEEE}&2021&U-Net+PatchGAN&Rigid,&Private, EndoVis17&Real,&Adversarial&Real&Segmentation(B$^{\rm a}$)\\
&&&Robotic&Synthetic&&&\\
\cite{kalia2021co}&2021&CoSegGAN&Robotic&EndoVis17, UCL ex-vivo,&Real,&Adversarial&Real&Segmentation(B$^{\rm a}$)\\
&&&& Private&Synthetic&&&\\
\cite{liu2021:MICCAI}&2021&SePIG&RObotic&EndoVis17,18&Real&UDA&Real&Segmentation(I$^{\rm a}$)\\
\cite{sahu2021:IJCAS}&2021&DNN+&Rigid,&Sim$^{\rm b}$, SimCholec80,&Real,&UDA$^{\rm c}$&Real,&Segmentation(B$^{\rm a}$)\\
&&Ternaus11&Robotic&EndoVis15&Synthetic&&Synthetic&\\
\cite{peng2021:arXiv}&2021&DeepLabv3+$^{\rm e}$,&Rigid,&UW Sinus, EndoVis17&Real,&Active- Learning&Real,&Segmentation(B$^{\rm a}$)\\
&&MobileNet&Robotic&&Synthetic&&Synthetic&\\
\cite{zhao2021one}&2021&MDAL&Robotic&EndoVis17,18,&Real&Meta-Learning&Real,&Segmentation(B$^{\rm a}$),\\
&&&&HKPWH$^{\rm f}$&&&Phantom& Tracking\\
\cite{zhao2021anchor:MIA}&2021&AOMA&Robotic&Davis16, EndoVis17,18,&Real&Meta-Learning&Real&Segmentation(B$^{\rm a}$),\\
&&&&HKPWH$^{\rm f}$&&&&Tracking\\

\hline

\end{tabular}
\begin{tablenotes}
$^{\rm a}$ B=Binary, P=Parts, I=Instance ;$^{\rm b}$ Simulated Data; $^{\rm c}$ Unsupervised Domain Adaptation; $^{\rm d}$ WS:Weakly Supervised SS:Semi-Supervised ; $^{\rm e}$ Modified Model; \\                             $^{\rm f}$ HKPWH:Hong Kong Prince of Wales Hospital; 
\end{tablenotes}
\end{threeparttable}
\end{center}
\end{minipage}
\end{sidewaystable*}

\subsubsection{Weak supervision}
Fully supervised semantic segmentation methods require a pixel-wise fully annotated data samples which is quite tedious and time-consuming apart from requiring technical expertise if the data related to the medical domain. This is the reason for the shortage of large labelled datasets in the endoscopic surgery domain in contrast to natural imagery; furthermore, those available are limited to shorter sequences. This limitation usually leads to over-fitting problems, severely restricts the model's generalizability and thus has motivated researchers to go for alternative techniques to build machine learning model, one of them being weakly supervised approaches. 

Semantic segmentation with weakly supervised learning (WSL) has been studied in several studies \citep{durand:CVPR,kim:CV,chang:CVPR}. Various kinds of weak supervisions have been performed in the literature such as pixel-level \citep{ahn:CVF}, scribbles \citep{lin:CVPR}, based on bounding boxes \citep{kervadec:MIDL} and points-based \citep{bearman:ECCV}. In the medical domain, WSL approaches have been explored for the detection of cancer regions \citep{jia2017:TMI,hwang2016:MICCAI}, prostate and brain lesion segmentation \citep{wang:MICCAI}, CT and MRI image registration \citep{blendowski:MIA} and so on. 

In the domain  of endoscopic tool localization, one study \citep{vardazaryan:Springer} uses WSL approach in which authors use image-level labels only without spatial annotations. In this work, the FCN architecture has been used to generate heat maps that provide for multi-class tool presence confidence values. In another work \citep{fuentes:IJCARS}, 'stripe' labels are defined from the contours of the instruments of the endoscopic image to train a DeepLabv3+ \citep{chen:2018} model for all three applications, i.e., binary, parts, and instance segmentation. The model has been tested on three different surgical datasets. The model still requires fully annotated segmentation masks to evaluate its performance. In a two-stage \citep{lee:IET} framework, weak labels were created by giving a multi-modal input to random walk and then a DCNN model is trained on the generated labels to perform binary instrument segmentation. \cite{sanchez:MICCAI} tackled the lack of data problem by using images with only image-level tool presence labels and train EfficientDet \citep{efficientdet:CVF} model to estimate the bounding box and segmentation mask for surgical instruments. In this comprehensive study, they validate the model on various fractions of annotated data in a semi-supervised way and show that the proposed approach obtains competitive performance compared to the fully supervised methods even with 1\% annotations. The problem with the approach may be the tool presence labels may produce undesirable outcome if more than one tool of the same is present in the image. In that case, adding temporal information may address the problem. 


\subsubsection{Semi-supervised learning in instrument segmentation}

With the development of techniques to exploit unlabelled data, semi-supervised learning has produced promising results \citep{van:ML}. \cite{yoon:Book} proposed a tool detection and tracking pipeline using pseudo-labels. But the limitation of that approach is that it relies on post-correction for the removal of incorrectly labeled data. An alternative is to use self-training and augmentation-driven consistency (STAC) \citep{sohn:arXiv} to avoid post-correction and produce high-quality pseudo-labels. 
Inspired by the technique of \cite{sohn:arXiv}, the authors in \cite{jiang:Springer} propose a semi-supervised teacher-student framework and address the post-correction limitation by adding an automatic filtering setup with a certain threshold. In this manner, the model produces confident pseudo-labels and a comparative evaluation shows that this semi-supervised approach attains improvements over supervised methods in mean average precision metric.  
A semi-supervised loss is computed by transferring predictions of unlabelled frames to near frames using optical flow in \cite{jin:MICCAI}. This approach relies heavily on the optical flow method. Another dual motion strategy \citep{zhao:MICCAI} addresses the problem holistically by predicting motion flows and performing joint learning to recover labels on a sparsely labeled dataset. In contrast to the general algorithms where adding more labelled data yields increasingly better segmentation results, authors in \cite{fu:Book} show that the opposite may also be true with a novel training strategy on mean Teacher model. The obtained results do improve the overall generalization capabilities of the model, but failures occur in a situation when blood is over the instruments or due to poor lighting conditions. 
A user-interactive mechanism for generating tool annotations was proposed in \cite{lejeune2018:MIA}. This study uses 2D point supervision approach to place dots on the expected instruments' locations in the image with only few assumptions and subsequently recovers the pixel-wise segmentation mask of the tool from the background by using a tracking approach. The proposed approach assumes that the object of interest is present in every frame. 

\subsubsection{Adversarial approaches}

CAS aims to enhance the efficiency and precision of surgical procedures to improve patient outcomes. Vision-based methods using deep learning have been instrumental in the implementation of CAS frameworks. However, the availability of annotated data is the main hurdle in developing a robust and generalized solution. To this end, various strategies have been adopted by researchers like crowd sourcing \citep{maier:MICCAI,maier2016:MICCAI}, generating synthetic data \citep{shorten:JBD}, active learning strategies \citep{peng2021:arXiv}, unsupervised training \citep{liu2020:MICCAI} or as previously discussed, using weakly or semi-supervised approaches. Furthermore, newer ways\footnote{SAGES Innovation Weekend - Surgical Video Annotation Conference 2020} of annotating surgical videos are also being explored by the research community.

\noindent\textbf{Surgical data augmentation and adversarial rendering}: Limited availability of training data in the medical domain has led towards increased use of GANs \citep{liu:NIPS}, which can generate real-like data without needing labelled samples. The lack of data problem for surgical tool segmentation has been approached by researchers by generating synthetic data \citep{shorten:JBD,lindgren:RSJ}. One technique is to use surgical simulators such as a 3D slicer \citep{kikinis:IIIT}, dV-trainer \citep{perrenot:SE}, AMBF \citep{munawar:ICRA}, the RobotiX mentor \citep{whittaker:JE} for image generation, but the acquired images lack the realistic artifacts or visual features found in real surgical settings. Another approach used in the literature is converting the working domain into fully synthetic ones by training models on synthetic data and the real converted to synthetic \citep{mahmood2018:TMI}. However, the problem with this approach is that the conversion process from real to synthetic loses much of the potential cues and details.

Traditional methods use morphological augmentation techniques which are not feasible for surgical domain since the surgical scenes contain rich textual and human tissue information. On the other hand, GAN-driven augmentation exhibits an inherent advantage over their traditional data augmentation counterpart in that GANs can enforce domain-specific features. The work presented in \cite{zisimopoulos:HTL} validates the effectiveness of using simulated data in order to train DL models. However, there is a considerable domain gap between simulated and real data which is addressed in works like in \cite{pfeiffer:MICCAI} using I2I translation techniques.

\begin{figure*}[t]
\centering
\includegraphics[width=5.5in]{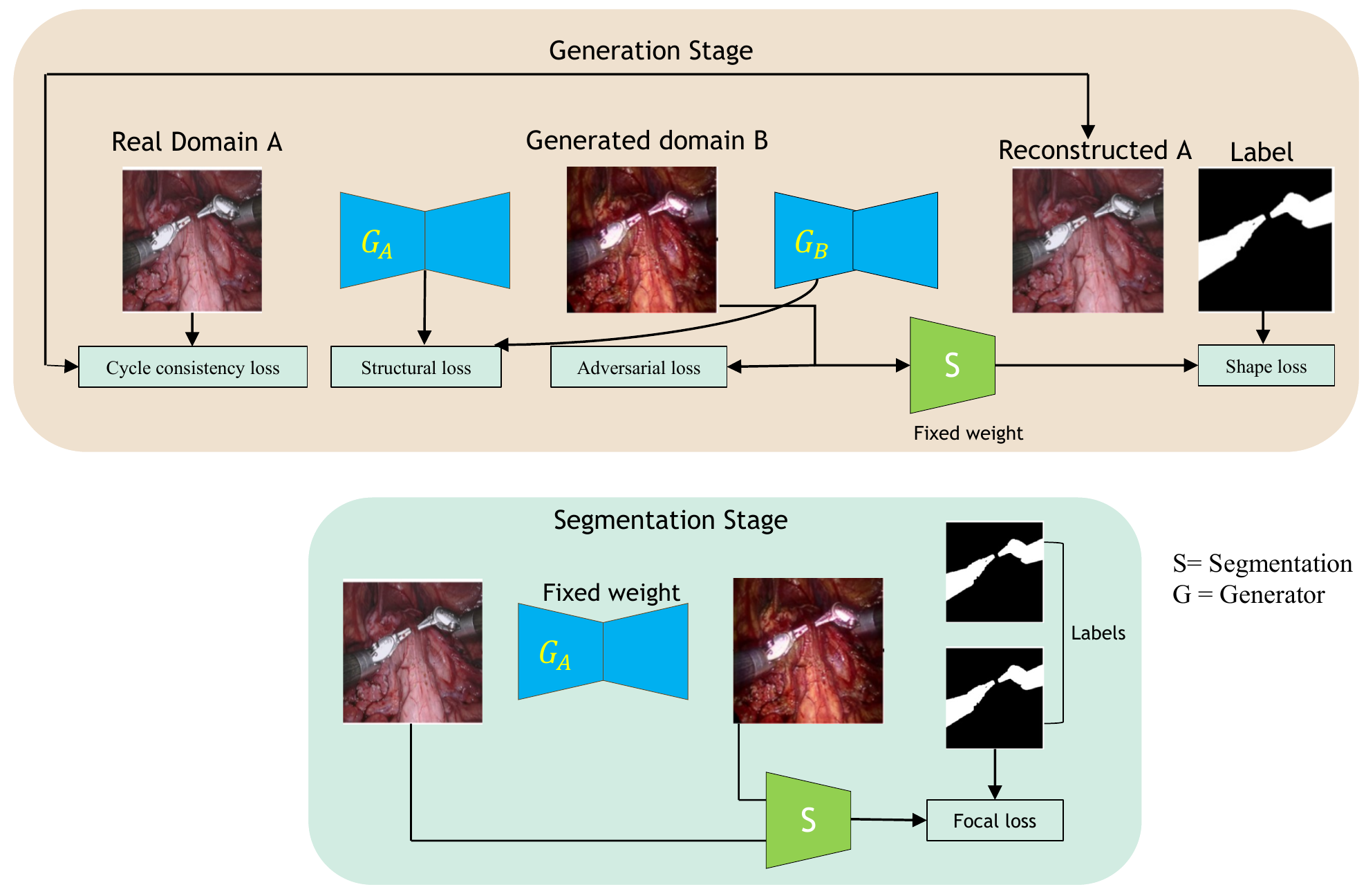}
\caption{Joint generation and segmentation strategy}
\label{fig6}
\end{figure*}

Having synthetic unlabelled data in abundance, the next step is how to effectively utilize its potential in tool localization. A self-supervised learning technique with GAN-based re-colorization has been proposed by \cite{ross:IJCAS} to effectively exploit the usefulness of unlabelled data for instrument segmentation. Multi-scale feature fusion together with adversarial loss is employed for instrument segmentation by \cite{islam:IEEE}. Instead of manual annotations, synthetic ones are rendered through a CAD system and robot kinematic model in \cite{pakhomov:RSJ}. Since the annotations are not precise enough to be used for supervised training, CycleGAN \citep{CycleGAN:CVPR} based techniques to learn mapping and make them cycle consistent are used. However, the model still performs poorly. To improve synthetic image generation and local style preservation, three GAN-based models \citep{su2021:sensors} were developed to generate realistic-looking tools while using real surgical background images. The first strategy deals with the conversion of the whole synthetic image dataset into  the real domain; the second approach works on only tool pixels, while the third combines both strategies. Models trained using the third strategy were validated using a U-Net architecture by training it on both real and synthetic data. Results indicate that the addition of synthetic data in training improves  performance. The study conducted in \cite{colleoni:IRAL} provides a comparative evaluation of models trained on real and synthetic data using different 12I architectures  for data generation. A synthetic dataset is obtained by performing I2I translation on simulated images and then blended with surgical background. However, tool pixels are processed without actual backgrounds which do not include challenging artifacts like reflections, motion blur or tool tissue interactions. Arguing that simple I2I approaches or computer simulations cannot reproduce challenging surgical scenes, \cite{ozawa2020:TF} generated a synthetic dataset by combining real laparoscopic videos and computer graphics generated images and then validated the technique on a modified U-Net architecture. LC-GAN \citep{lin2020:IROS} tries to address the lack of visual effects problem by cross-domain image synthesis between cadaver(source) images to live (real) surgery images. These approaches either require the availability of paired training data or four adversarial networks which increases the training time.   

GAN-based approaches have produced promising results in data generation, but they suffer from few problems like the presence of undesirable artifacts in their output \citep{colleoni:IRAL} or require large amounts of training time. In order to address the training time considerations, an unpaired I2I framework is proposed by \cite{zhang2021:IEEE} with a reduced number of generators and discriminators in comparison to architectures such as DualGAN. Tackling the artifacts problem, a joint generation and segmentation strategy was developed by \cite{kalia2021co} for better domain generalization from labelled to unlabeled data in instrument segmentation. The proposed framework is summarised in Fig. \ref{fig6}.

\textbf{Domain adaptation and generalization}: Domain shift problems have been a challenge for DL models in surgical scenes. Supervised methods fail to generalize well on cross domain-datasets. Some methods use image I2I for addressing domain gap problem. To this end, \cite{pfeiffer:MICCAI} uses CycleGAN for image-level style transfer, while work proposed by \citep{sahu2021:IJCAS} uses Endo-Sim2Real \citep{sahu2020:MICCAI} consistency based framework for end-to-end joint learning from simulated and real domains.  However, these techniques focus on appearance-based image translation and accuracy rests on the image translation quality. These methods perform well on binary segmentation but under-perform on multi-instrument scenarios. Semantic-prototype interaction graph (SePIG) \citep{liu2021:MICCAI} architecture proposes better feature alignment technique between source and target domain for instrument type segmentation. 

\subsection{Instrument detection}
This section highlights the instrument bounding box detection problem and recognition of instruments. The overview of instrument detection SOTA is provided in Table \ref{tab5}. Tool presence detection and localization can also prove to be handy in several surgical applications like video indexing \citep{endonet:TMI} on a surgical dataset and detecting any upcoming complications by identifying tool in a scene that should not appear. Tool detection algorithms normally take an input video frame of a surgical procedure and produce output bounding box coordinates for the tool location. Instrument detection approaches have been traditionally designed using hand-crafted features or through learned visual data-driven features. The limitation with hand-crafted features are that they do not consider the relationship between different tools, while learned feature based methods do not provide better accuracy in cases of occlusions, blood or specular reflections. 

In some of the earlier works, a multi-task network EndoNet \citep{endonet:TMI} was proposed for tool presence detection and phase recognition by transfer learning given the scarce annotated samples. The limitation is that features lack temporal context. 

Surgical tool detection techniques based on deep learning can be subdivided into two categories namely one-stage and two-stage methods. In a one-stage framework, \cite{choi2017surgical} proposed a modified YOLO architecture to directly regress tool bounding box coordinates. The model achieved faster speed but at the expense of accuracy.  Another single-stage detector is proposed by \cite{kurmann2017simultaneous} with modified CNN architecture with the assumption that the maximum number of instruments in a frame is known. To reduce computation time and enhance inference speed, a novel CNN architecture with a ghost module for generating feature maps while using the \emph{Mish} activation function from CSPNet \citep{wang2020:CVF} was proposed by \cite{yang2021:MIUA}. The results from the framework show encouraging performance with 100\% mAP on EndoVis dataset while 91.6\% mAP on the Cholec80-locations dataset.  In a two-stage technique,\citep{sarikaya2017:TMI} proposed the first approach that incorporates DNNs for tools detection and localization in robot-assisted surgery. The network uses two deep models that take an RGB video frame and its corresponding optical flow information. In this work, they also introduced the ATLAS Dione dataset, the first public set of data on robot-assisted surgery videos with tool annotations. However, the network inference speed is quite low (less than 10 FPS). Another region-based technique based on Faster-RCNN was proposed by \cite{jin2018tool:WACV} with the introduction of a new tool detection dataset named m2cai16-tool-locations. The proposed model is designed for multi-class detection with the clipper having the highest detection accuracy while the irragator has the lowest due to its generic shape and less frequent presence in the data. A common limitation among single and multi-stage methods is a trade-off between accuracy and speed. Following the Mask-RCNN based architectures, authors in \citep{kletz2020:ICMM,kletz:CBMI} present comprehensive experiments on classification and pixel-wise segmentation on 11 different surgical instrument types. 
These classical object detection pipelines have fixed receptive field which makes it difficult to locate multi-scale objects in an image. Another approach in the literature disregards instrument types present in the image and just focuses on tool detection in a Lapgyn4 dataset \citep{leibetseder2018:ACM}.  Most of the other SOTA methods focus on aggregating multi-stage features by designing a pyramid structure such as Scale-Transferrable Detection Network (STDN) \citep{zhou2018:CVPR} and Single- Shot Refinement Neural Network (RefineDet) \citep{zhang2018:CVPR}. However, these methods may lead to class imbalance problem due to having fix box size and aspect ratio. 
The class imbalance problem among the dataset has been addressed by work in \cite{sahu2017:IJCARS} which used a re-sampling strategy to create balanced training data based on the tool occurrences frequency while \cite{alshirbaji2018:BM} employed weighted loss along with re-sampling.

 \begin{sidewaystable*}[!htbp]
\tabcolsep7.5pt
\footnotesize
\begin{minipage}{\textheight}
\caption{Review of Instrument detection SOTA}\label{tab5}
\begin{center}
    
\begin{threeparttable}
\begin{tabular}{@{}llllllllll@{}}\toprule

\multicolumn {5}{c}{\textbf{Training}} &&&\multicolumn{1}{c}{\textbf{Test}} \\\cmidrule(r){3-7}\cmidrule(l){8-8}

Ref.&Year&Architecture & Tool & Dataset & Data & Technique& Data&Task&classes\\
\hline
\cite{endonet:TMI}&2016&EndoNet&Rigid&Cholec80,&Real&Supervised&Real&D,C,P$^{\rm a}$&7\\
&&&&EndoVis15&&&&&\\
\cite{hu2017agnet:DLMIA}&2017&AGNet&Rigid&m2cai16-tool&Real&Supervised&Real&D,C$^{\rm a}$&7\\
\cite{mishra2017:CVPR}&2017&CNN+LSTM&Rigid&m2ccai16-tool&Real&Supervised&real&D,C$^{\rm a}$&8\\
\cite{kurmann2017simultaneous}&2017&CNN&Rigid&RMIT,&Real&Supervised&Real&D$^{\rm a}$&4\\
&&&&EndoVis15&&&&&\\
\cite{sarikaya2017:TMI}&2017&CNN+RPN$^{\rm c}$&Robotic&ATLAS Dione&Phantom&Supervised&Phantom&D$^{\rm a}$&NA$^{\rm f}$\\
\cite{jin2018tool:WACV}&2018&Faster RCNN+&Rigid&m2cai16-tool-locations&Real&Supervised&Real&D,C$^{\rm a}$&7\\
&&RPN&&&&&&&\\
\cite{vardazaryan:Springer}&2018&FCN&Rigid&Cholec80&Real&WS$^{\rm b}$&Real&D,C$^{\rm a}$&7\\
\cite{colleoni2019:RAL}&2019&3D FCNN&Robotic,&EndoVis15,&Real,&Supervised&Real,&D$^{\rm a}$&NA$^{\rm f}$\\
&&&Rigid&UCL dVRK&Sim&&Sim&&\\
\cite{kletz:CBMI}&2019&Mask RCNN&Rigid,&Custom&Real&Supervised&Real&D,C$^{\rm a}$&11\\
&&&Robotic&&&&&&\\
\cite{wang2019graph:MICCAI}&2019&ST-GCN&Rigid&m2cai16-tool,&real&Supervised&Real&D$^{\rm a}$&7\\
&&&&Cholec80&&&&&\\
\cite{yu2020massd:CBM}&2020&MASSD&RObotic&ATLAS Dione&Phantom&Supervised&Phantom&D$^{\rm a}$&NA$^{\rm f}$\\
\cite{shi2020real:Access}&2020&CDM +RDM$^{\rm d}$&Robotic,&ATLAS Dione,&Real,&Supervised&Real,&D,C$^{\rm a}$&7\\
&&&Rigid& EndoVis15&Phantom&&Phantom&&\\
&&&&Cholec80-locations&&&&&\\
\cite{liu2020anchor:Access}&2020&CNN+&Rigid,&ATLAS Dione,&Real,&Supervised&Real,&D$^{\rm a}$&NA$^{\rm f}$\\
&&Bourglass&Robotic&EndoVis15&Phantom&&Phantom&&\\
\cite{zhang2020surgical:Access}&2020&Faster RCNN +&Rigid&AJU-Set,&Real&Supervised&real&D,C$^{\rm a}$&7\\
&&RPN$^{\rm c}$&&m2cai16- tool- locations&&&&&\\
\cite{kletz2020:ICMM}&2020&MaskRCNN+&Rigid,&Private&Real&Supervised&Real&D$^{\rm a}$&7\\
&&FPN$^{\rm e}$&Flexible&&&&&&\\
\cite{yoon:Book}&2020&Faster, &Robotic&Private&Real&SS$^{\rm b}$&Real&D,C$^{\rm a}$&14\\
&&Cascade RCNN&&&&&&&\\
\cite{kondo2021:CMBBEI}&2021&CNN+ &Rigid&Cholec80&Real&Supervised&Real&D,C$^{\rm a}$&7\\
&&Transformer&&&&&&&\\
\cite{yang2021:MIUA}&2021&GhostNet+&Rigid,&EndoVis15,&Real&Supervised&Real&D,C$^{\rm a}$&7\\
&&Yolov3&Robotic&Cholec-80- Locations&&&&&\\
\cite{alshirbaji2021:BMSC}&2021&CNN+LSTM&Rigid&Cholec80&Real&Supervised&Real&D,C$^{\rm a}$&7\\
\cite{namazi2021:SE}&2021&RCNN&Rigid&M2CAI16, Cholec-80&Real&Supervised&Real&D,C$^{\rm a}$&7\\
\cite{jiang:Springer}&2021&Faster R-CNN&Rigid&CaDTD&Real&SS$^{\rm b}$&Real&D$^{\rm a}$&12\\
\cite{teevno2022}&2022&Teacher-&Rigid&m2cai16&Real&SS$^{\rm b}$&Real&D$^{\rm a}$&7\\
&&Student&&&&&&&\\
\bottomrule
\end{tabular}

\begin{tablenotes}
$^{\rm a}$ D=Detection, C=Classification P=Phase Recognition ; $^{\rm b}$ SS=Semi-Supervised, WS=Weakly Supervised;   $^{\rm c}$ Region Proposal Network; $^{\rm e}$ CDM= Coarse Detection Module, RDM= Refined Detection Module; $^{\rm e}$ Feature Pyramid Network; $^{\rm f}$  Not Available\\ 

\end{tablenotes}
\end{threeparttable}
\end{center}
\end{minipage}
\end{sidewaystable*}

The trade-off between performance and inference time has been addressed by \cite{zhang2020surgical:Access} by proposing modulated anchoring framework instead of fixed anchors. The results indicate a significant performance enhancement in tool detection but authors did not mention the inference time. \cite{zhao2019:HTL} proposed a dual CNN framework for tool detection. One network produces heatmaps while the other takes heatmaps as input and produces bounding boxes. The results indicate better performance on accuracy and speed but could not use end-to-end training. An end-to-end training was achieved in a later work using anchor-free CNN architecture for tool detection \citep{liu2020anchor:Access}.

In an effort to address the problems of artifacts, attention mechanisms have been proposed in SOTA. In this regard, AGNet \citep{hu2017agnet:DLMIA} uses a two-step approach for tool detection. The model predicts visual attention maps through a global prediction network and the local networks output prediction for each tool. 
Inspired by RefineDet \citep{zhang2018:CVPR}, authors in \cite{shi2020real:Access} proposed  a single-stage tool detection technique by combining attention module with a light-weight network for tool detection. The model achieved promising results with 100\% mAP and 55.5 FPS  on EndoVis challenge. Using attention mechanism for feature extraction, \cite{yu2020massd:CBM} proposed several feature fusion techniques based on multi-level and semantic information for tool detection. Though the model is light-weight and single-stage, results indicate competitive performance with two-stage approaches. 

In order to address the lack of annotated datasets problem, \cite{vardazaryan2018weakly} use image-level labels to train a pre-trained FCN model and re-train on the Cholec80 dataset getting mAP of 87.2\% for binary tool presence. In another work, Teacher-Student joint learning was proposed by \cite{teevno2022} to detect surgical tools. The method further proposed a novel margin-based distance loss to segregate effectively segregate foreground from the background. 

Since CNN-based models extract only a high level information from surgical images, they lack temporal connection between multiple instruments present in the image leading to lesser detection performance. In this context, \cite{mishra2017:CVPR} trained LSTM network along with CNN to incorporate temporal information for better tool detection. LapTool-Net \citep{namazi2021:SE} used Gated recurrent network for the detection of seven different tools. 
\cite{chen2018:Springer} proposed 3D CNN to learn spatio-temporal cues from short surgical videos. 3D FCNN architecture similar to U-Net with skip connections was proposed for surgical instrument joint detection by \cite{colleoni2019:RAL}. Tool usage detection framework using combined CNN and RNN was proposed by \cite{al2018:MIA}. Instead of training CNN-RNN networks end-to-end, authors introduce weak classifiers to supervise CNN training as per RNN output. \cite{wang2019graph:MICCAI} explored graph convolutional network to incorporate spatio-temporal information across successive frames for tool presence detection. The model used labelled frames and unlabelled adjacent frames for training resulting in significant performance improvement. Many of these types of approaches either include temporal dependencies from short videos as most of the datasets are sparsely annotated or consider long-term video information. Building on the premise that short-term sequences may help in detection because surgeons may use the specific tool in the specific phase and long-term temporal dependencies for tool detection in cases when misclassifications can be revised later, tool presence detection was approached as spatio-temporal problem by \cite{alshirbaji2021:BMSC}. They used a cascade of two LSTMs, one for modeling short-term while other for long-term temporal information. LapFormer \citep{kondo2021:CMBBEI} added transformer module with CNN to incorporate temporal information for surgical tool detection. The results show the superior performance of the transformer over methods based on LSTM. Ensemble learning offers a way to add the strength of multiple models to boost performance. To that end, GoogleNet and VGGNet for tool detection on M2CAI-tool detection dataset \citep{wang2017:ISBI} were trained for tool detection.

\subsection{Instrument tracking}

Pose estimation of an instrument at any given time during the surgical process is of great importance since it gives a clue as to how much is the distance between the tool and any critical structure. Instrument pose can also help in automating surgical skills or performing skill assessment. In order to estimate tool pose, tool tracking throughout the surgical procedure becomes essential. A summary of tool tracking algorithms can be found in Table \ref{tab6}. Early approaches rely on surgical tool information gathered through color, texture or geometric constraints \citep{zhou2014:ACE}, gradients \citep{sznitman2012:MICCAI}, aggregation of those features \citep{reiter:RSJ}, while other approaches use markers mounted on the instruments \citep{loukas2013:IJMRCAS,zhang:IJCARS}, superpixel-based tracking \citep{yang2014:TIP} or combined optical and marker-based tracking systems \cite{zhou2017:JBO,ou2020:ICVRV}. Although the presence of markers makes tracking pipeline more robust and simple, it is quite cumbersome and pose sterilization problems. Furthermore, the applicability of optical trackers is severally limited by the field of view, line of sight and occlusion problems. In contrast to markers or sensors-based tool tracking, image-based methods are non-invasive and have the ability to produce the tool pose directly on the surgeon's viewing screen. \\
Machine Learning-based approaches have used visual features such as edge features \citep{pezzementi:IEEE} or fast coroners \citep{reiter:RSJ} to train appearance models for tracking. The study presented in \citep{reiter2012:MICCAI} uses natural tool landmarks for feature-based tracking. Change of appearance problem in tool tracking was addreesed by \cite{li2014:MICCAI} using online learning technique. Other studies \citep{du2016:IJCARS} have used Scale-invariant feature transforms SIFT, or Histogram of gradients (HoG) for feature based tracking along with some classifiers like support vector machines (SVM) or random forest. However, motion artifacts, the presence of blood, smoke, and light reflections limit the applicability of feature descriptors. The technique used in \cite{du2016:IJCARS} required manual initialization of tracker which is not suitable for clinical translation. Recently, vision-based and marker-free methods with the aid of deep learning architectures have been explored by several researchers and have shown promising results. 

Data from a study conducted in \cite{bodenstedt2018comparative} shows that tracking methods perform well on ex-vivo datasets while significant degradation in performance is observed in in-vivo. Generally, single object tracking techniques did not do well on multiple instrument tracking, out-of-view cases and occlusion situations. Other approaches \citep{ye2016:MICCAI} used in the literature employ robotic kinematic information and 3D CAD models which may invariably restrict their application in clinical translation. These methods perform poorly in tool occlusion scenarios while their real time implementation is prohibitively expensive. Specialized near-infrared optical systems have also been used for multi-instrument tracking in surgery \citep{cai2016:CAS}. 

\begin{sidewaystable*}[!htbp]
\tabcolsep7.5pt
\footnotesize
\begin{minipage}{\textheight}
\caption{Review of Instrument Tracking SOTA}\label{tab6}
\begin{center}
\begin{threeparttable}
\begin{tabular}{@{}lllllllll@{}}\toprule

\multicolumn {5}{c}{\textbf{Training}} &&&\multicolumn{1}{c}{\textbf{Test}} \\\cmidrule(r){3-7}\cmidrule(l){8-8}

\textbf{Ref.}&    \textbf{Year}& \textbf{Architecture} & \textbf{Tool} & \textbf{Dataset} & \textbf{Data} & \textbf{Technique} & \textbf{Data} & \textbf{Application Task}\\
\hline
\cite{allan2015:MICCAI}&2015&DT$^{\rm a}$,OF$^{\rm a}$ &Robotic&Self&Real&Features&Real&Tracking, 3D pose\\
\cite{kurmann2017simultaneous}&2017&CNN&Rigid&in-vivo RM,&Real&Dtection&Real&DT$^{\rm b}$\\
&&&&EndoVis15&&&&\\
\cite{zhao2017tracking}&2017&CNN&Rigid&Private&Real&Detection&Real&tracking\\
\cite{sarikaya2017:TMI}&2017&CNN+RPN,&Robotic&ATLAS Dione&Phantom&Detection&Phantom&DTL$^{\rm b}$\\
&&Fast RCNN&&&&&&\\
\cite{lejeune2018iterative}&2018&U-Net&Rigid&BRATS,& Real&Features&Real&Tracking,\\
&&&&EndoVis15, Cochlea&&&&Segmentation\\
\cite{du2018articulated}&2018&FCN&Rigid&RMIT, EndoVis15&Real&Detection-Regression&Real&Tracking, Pose\\
\cite{nwoye2019:IJCARS}&2019&FCN+&Rigid&Cholec80&Real&Detection&Real&DTL$^{\rm b}$ \\ 
&&ConvLSTM&&&&&&\\
\cite{zhao2019real:CAS}&2019&CNN+STN+&Rigid,&Self+&Synthetic,&Detection&Synthetic,&Tracking\\
&&STC&Robotic&Public&Real&&Real&\\
\cite{qiu2019:HTL}&2019&RT-MDNet&Robotic,&m2cai16-tool,&Real&Detection&Real&Tracking\\
&&&Rigid&STT$^{\rm f}$&&&&\\
\cite{du2019patch}&2019&PAWSS&Rigid&OTB, VOT,&Real&Seg$^{\rm c}$&Real&Tracking\\
&&&&EndoVis15&&&&\\
\cite{lee2020:JCM}&2020&MaskRCNN,&Robotic&BABA$^{\rm d}$,&Real&Seg$^{\rm c}$,Detection&Real&Tracking,SA$^{\rm b}$ \\
&&Deep SORT&&ST-ReID,Others$^{\rm e}$&&&&\\
\cite{zhang:IJCARS}&2020&LinkNet&Rigid&m2cai16-tool&Real&Seg$^{\rm c}$&Real&Tracking\\
\cite{islam2021:MIA}&2021&ST-MTL&Robotic&EndoVis17&Real&Seg$^{\rm c}$&Real&Tracking,\\
&&&&&&&&Seg$^{\rm c}$\\

\bottomrule
\end{tabular}
\begin{tablenotes}
$^{\rm a}$ DT= Decision Tree, OF= Optical Flow ; $^{\rm b}$ DTL= Detection, Tracking, Localization, SA=Skill Assessment; $^{\rm c}$ Segmentation; $^{\rm d}$ BABA Training Model;  $^{\rm e}$ ST-ReID:Patients with thyroid cancer, EndoVis17; $^{\rm f}$ Surgical Tool Tracking Dataset                           
\end{tablenotes}
\end{threeparttable}
\end{center}
\end{minipage}
\end{sidewaystable*}

\subsubsection{Tracking by detection} 
Generally, tracking by detection methods train a classifier to separate the region of interest from the background and then keep on updating with new information in each frame. The reliability of these methods may be compromised if some of the samples are incorrectly labelled. Furthermore, the substantial challenge for tracking by detection methods is that the bounding box not only contains the object of interest but rather a considerable portion of background too. This has an adverse effect on the model training since background portion keeps on hanging in different frames. One way to address the problem is to use a dynamic weight-assigning strategy where different pixels are assigned a different weight, based on its spatial location or appearance similarity to suppress the background pixels and highlight foreground \citep{lee:CVPR,he2013:CVPR}. This technique has been incorporated in \cite{xie2015:ICV} to assign variable weights to image patches. In another study called Patch-based adaptive  weighting with segmentation and scale (PAWSS) \citep{du2019:MIA}, a color-based segmentation stage has been incorporated for an improved weighting strategy. In this framework, the object bounding box is uniformly distributed into non-overlapping image patches. However, the approach is only limited to single object tracking and assumes that bounding box covers most of the instrument region.

An articulated surgical instrument consists of two parts: Shaft and end-effector. It is often desirable that a model jointly track both parts for a better and accurate pose estimation of the instrument. To this end, \cite{zhao2017:CAS} proposed part-based tracking of both parts of the surgical instrument. The approach uses line detection for shaft portion which may provide erroneous results if it contains blood. Furthermore, if the angle of imaging is changed, the shaft localization would throw a location error. In another study \citep{du2018:IEEE}, a probabilistic map instrument joint detection framework was proposed to improve the tracking robustness. 

Multiple tool tracking frameworks have also been investigated in the literature. One possibility to perform multiple instrument tracking can be to deploy multiple object trackers for each instrument. But the limitations will outweigh the benefits such that these will be quite computationally expensive, and would result in drift accumulation or occlusions. In other methods for multiple instrument tracking for instance, WSL study conducted by \cite{nwoye2019:IJCARS} uses frame-level labels to detect tool tip and track multiple instruments resulting in 12.6\% improvement in SOTA. The proposed approach uses ConvLSTM to consider temporal coherence for tracking. In this approach, however, when multiple tools of the same class are present in the image, then mere tool presence label would not suffice. Also the model fails to track tools if only 1/5th of its portion is visible in the frame, while the shaft is mis-detected because several tools have a similar shaft. \cite{chen2017:CAC} performed tool tracking by generating their own labels through the line segment detector and then training CNN on those labels to detect and track tool tip.  

In order to solve the occlusion and boundary problems of tool detection, spatial transformer network (STN) is used for efficient tool localization while spatio-temporal context (STC) performs frame-by-frame tracking \citep{zhao2019real:CAS}. CNN and STN detection rate being slow (2.5s), the proposed method uses STC to get spatio-temporal information for the tool tip in order to track its position in real-time achieving 48 FPS on i3 machine. A two-stage framework \citep{lee2020:JCM} consisting of instance segmentation to address occlusion problems and  deep simple online real-time tracker (DeepSORT) \citep{khalid:JAMA} is used for frame-wise tool tip tracking was proposed. To avoid tool tracking problem because of re-appearance into the scene, it integrates re-identification module to deepSORT thus maintaining instrument identity in long sequences. The limitation of this could be the accumulation of errors in long video sequences because of multiple algorithms used. With a maximum number of instruments in an image known and to avoid complexities of two-stage frameworks, CNN based multi-instrument recognition and parts' 2D points estimation is proposed by \cite{kurmann2017simultaneous}. While the approach outperforms other tracking methods, joint occlusions hinder 2D point estimation. Also the inference time in cases of high input image size is around 6 FPS. 

Real-time segmentation and tracking have been proposed in the work \citep{laina2017concurrent} which exploits the inter-dependency between detection and segmentation by jointly implementing both tasks using CNN. In an effort to make tracking real-time, coarse to fine light-weight cascaded CNN architecture is proposed by  \cite{zhao2019:JoE}. \emph{Coarse CNN} locates the tool while \emph{fine} acts as a regression network for tool tip tracking. The method achieves an FPS of 25. Recently, it has been shown that Fast-RCNN can localize surgical instrument with an excellent precision \citep{du2018:TMI}, however the problem with this approach is that it is computationally expensive. YOLO \citep{yolov3:arXiv} offers a viable alternative to that technique though. Authors in \cite{li2021:ASME} use Yolov3 to detect surgical tool tip, camera parameters for tip location and visual tracking space vector for multiple tool tracking. Efficiency of the tracking by detection techniques rely upon the detection step. The detection stage relies in turn on the availability of sufficient size of dataset, however datasets with sequential information are not easily available.  In this regards, a deep learning based approach was proposed on self-developed sequential dataset called surgical tool tracking (STT) \citep{qiu2019:HTL}. The proposed technique uses real time multi-domain CNN (RT-MDNet) \citep{jung2018:ECCV} which can be trained for each type of instrument. However,  the model result in terms of FPS is only 14. Optical flow was used for short-term tracking by \cite{garcia2016real}. A multi-object tracking framework proposed in \citep{robu2020:BM} could only track the instruments in 80\% of the tool trajectory. 
Tracking by detection approaches generally have higher inference time and low accuracy since the bounding can not fit well into the instrument.

\subsubsection{Tracking by segmentation}
Surgical tool tracking problem has also been approached in the literature by first segmenting the tool. To that end, \cite{zhang:IJCARS} proposed LinkNet \citep{linknet:VCIP} to extract the surgical tool shaft and end-effector and then track the instrument frame-by-frame using a tracking point. While the method uses light-weight segmentation network to provide real-time speed, its output accuracy greatly hinges on the performance of segmentation model. Authors in \cite{amini2016:MISAT} use traditional computer vision techniques for tool segmentation and then perform tracking using cameramen robot \citep{mirbagheri:SI}. However, this technique is not accurate since traditional CV methods are not efficient in the surgical domain. A multi-task learning (MTL) framework based on attention mechanism and task-specific saliency prediction was proposed for instrument segmentation, scanpath prediction and camera prediction \citep{islam:MICCAI}. The major limitation of the work was that the temporal information was taken into account.  Therefore, spatio-temporal multi-task learning (ST-MTL) framework was proposed by adding ConvLSTM++ to the framework in \citep{islam:MICCAI}  to efficiently predict surgical tool scanpath \cite{islam2021:MIA}. In an effort to develop a tracking framework for long-term trajectories \cite{robu2021towards} inspired by the work in \citep{yang2014:TIP} propose its modification by adding a binary segmentation step to get refined targets for tracking as well as reducing the impact of background pixels in the bounding box. The proposed model is well suited for long duration trajectories and uses multiple objects tracking management module. Tracking by segmentation methods are generally faster and accurate in comparison to tracking by detection. 

\subsubsection{Tracking in 3D}
Visual tracking methods have produced excellent performance in tracking and pose estimation of surgical instruments. However, these methods can only generate 2D pose from images. For a 3D tool pose and orientation detection, specialized fiducial markers embedded in the instruments could be one alternative but it is not clinically feasible \citep{zhao2016:Patents}. Gradient and color features have been combined to form a 3D tool tracking pipeline for robotic instruments in \citep{reiter2014:IJRR} and as a brute force matching in the virtually generated templates \citep{reiter2012:CASR}. Particle filter tracking scheme is used to track geometrically rendered tools by using robot forward kinematics and bayesian state estimation \citep{hao2018:RSJ}.  These methods can be implemented in real time with the aid of GPU, but it requires robot kinematic data which limits method's applicability for non-robotic tools. Large-scale region-based features are fused with low level features to perform tracking using optical flow in \cite{allan2015:MICCAI}. The proposed approach is able to yield six degrees of freedom of instrument with respect to the camera but only achieves tracking accuracy of 4.09nm. 
In an effort to better track the instrument in 3D, a multi-constraint energy maximization strategy is proposed by authors in \cite{allan2014:IPCAI}. They use region based contours learned through random forest for tool localization, while a stereo vision constraint for the depth information and temporal cues have been incorporated through Kalman filter. Results indicate decrease in the instrument pose estimation error in all three axes compared to SOTA. Study conducted in \cite{li2021:ASME} uses single-hole camera intrinsic parameters to transfer the detected tip location coordinates (3D) into camera coordinate system. 

\subsection{Depth perception}
In a MIS navigation setup, surgeons have to face mental fatigue and burden of associating data from pre- and intra- operative phases. In this situation, depth perception of intra-operative surgical scenes can be useful in doing pre- and intra- operative image registration. Depth perception of endoscopic data during intra-operative phase may greatly help in understanding surgical scenes resulting in improved patient-care. It may also help in imparting better surgical training to the attendees. Depth estimation can be achieved through various types of endoscopic data like structured light endoscopes \citep{lin2017:MICCAI}, monoscopic \citep{ozyoruk2021:MIA} or stereo endoscopes \citep{ye2016:MICCAI}. Structured light endoscopy uses known and projected light to reconstruct 3D image of tissues. Main advantage being that there is no limitation of texture information but requires specialised hardware. A review of various techniques used in depth perception is discussed in preceding sections while the precise information is given in Table \ref{tab7}.

\subsubsection{Monocular depth estimation}
Measuring depth from smonocular endoscopes has been a great challenge because it requires camera pose parameters, which in case of endoscopic camera is difficult to acquire. Estimating depth from monocular images have been approached by using SLAM, Structure from Motion (SfM), Shape from Shading (SfS) as well as Deep learning being combined with one of these techniques. SfM exploits image sequences obtained by camera at different instances to reconstruct 3D while SfS utilises light for the task. SfM is used in \cite{ma2019real:MICCAI} to generate depth maps from colonoscopy images to train the network into depth estimation mode. However, inherent limitation of SfM  to generate good quality depth maps from texture-less surfaces restricts its applicability. Combination of SfM and SfS is used for 3D reconstruction in \citep{zhao2016:MICCAI}. 

Other studies have explored the feasibility of synthetic data in generating depth maps and training adversarial networks for depth estimation. Synthetic data was generated by the blender in order to be used to train adversarial network in \cite{mahmood2018:MIA}. While computer-generated data is not realistic, indigo carmine (IC) blue dye is used for 3D reconstruction to address that problem \citep{widya2021:IP}. However, IC is not readily available to use. Self-supervisory signals obtained from sparse SfM signals are used to implement monocular depth estimation \citep{liu2019:TMI}. Similar strategy but with dense prediction in a probabilistic manner to highlight poorly lit regions is adopted in \cite{liu2020depth:MICCAI}. \cite{rau2019:IJCASR} uses pix-to-pix to estimate depth from a colonoscopy image by using a synthetic image and its ground truth. However, the approach does not use depth maps from real data to be included in GAN loss to bridge the domain gap. This problem has been addressed by \cite{cheng2021:MICCAI} which uses synthetic data to train GAN based depth estimation network along with exploiting unlabelled real data by using temporal information. The major limitation of this approach is being computationally expensive since it uses many GAN models. While most adversarial approaches take synthetic data and convert them into real-like images for model training, the reverse of that was proposed in \cite{mahmood2018:TMI} to learn a transformer using synthetic data to convert real images into synthetic representations. The depth estimation model was trained on these domain-adapted images with a self-regularization network to preserve clinically useful features while removing any patient-specific data. 

Several works have proposed unsupervised or self-supervised approaches for depth estimation since it is quite hectic to annotate endoscopic data with depth labels. In this perspective,  RoboDepth \citep{li2021:ICRA} model uses kinematic information from the surgical robot and tool tip segmentation mask generated by U-Net architecture to estimate scale-aware depth from monocular images. The model achieves an inference time of 20ms. \cite{turan2018:RSJ} used a different kind of unsupervised approach for depth estimation. The proposed network uses view synthesis cues obtained from multiple images taken at different camera poses as a supervisory signal for model training. Results show improved performance of depth estimation with only some errors on edges and lower textured regions. However, the major limitation of the method is that it requires a camera calibration matrix which is infeasible to obtain for practical purposes since datasets from multiple hospitals are to be used for model training. Another unsupervised approach uses stereo datasets for training a monocular depth estimation module as an error minimization problem \citep{xi2021:CMPBM}. \cite{recasens2021endo} train a self-supervised model that generates pseudo-RGBD frames for camera pose tracking and 3D scene reconstruction.   Unsupervised monocular depth estimation was also investigated in \cite{godard2017:CVPR} using novel training loss to enforce left-right consistency. \citep{sharan2020:BE} further improves upon the idea by using SOTA Monodepth model \citep{godard2019:CVF} and minimum re-projection loss and auto-masking. 

\begin{sidewaystable*}[!htbp]
\tabcolsep7.5pt
\footnotesize
\begin{minipage}{\textheight}
\caption{Review of Surgical Depth Perception SOTA}\label{tab7}
\begin{center}
\begin{threeparttable}
\begin{tabular}{@{}lllllllll@{}}\toprule

\multicolumn{4}{c}{\textbf{Training}} &&&\multicolumn{1}{c}{\textbf{Test}} \\\cmidrule(r){3-6}\cmidrule(l){7-7}

\textbf{Ref.}&    \textbf{Year}& \textbf{Architecture} & \textbf{Dataset} & \textbf{Data} & \textbf{Technique} & \textbf{Data} & \textbf{Application Task}\\
\hline
\cite{zhao2016:MICCAI}&2016&Groupwise&Private&Phantom,Real,&SfM&Real,Phantom,&3D Reconstruction\\
&&Registration&&Synthetic&&Synthetic&\\
\cite{mahmoud2017:arXiv}&2017&ORB-SLAM&NA&NA&SLAM&NA&Depth,\\
&&&&&&&3D reconstruction\\
\cite{mahmood2018:TMI}&2018&GANs&Private,&Real,&Adversarial,&Synthetic&Depth\\
&&&Public&Phantom&Monocular&Synthetic&\\
\cite{mahmood2018:MIA}&2018&FCN&Private,&Real,&Monocular&Real,&Depth\\
&&&EndoVis15&Synthetic&&Synthetic&\\
\cite{qiu2018:CVPR}&2018&ORBSLAM&NA&NA&Monocular SLAM&NA&Depth\\
\cite{mahmoud2018:TMI}&2018&ORB-SLAM$^{\rm a}$&Hamyln&Real&Monocular&Real&Depth,\\
&&&Private&&&&Tissue Tracking\\
\cite{turan2018:RSJ}&2018&Depth-CNN&Private&Real& Monocular&Real&Depth,\\
&&&&&Unsupervised&&Pose\\
\cite{chen2019slam:arXiv}&2019&cGAN&Private&Synthetic&Adversarial&Synthetic&Depth\\
&&&&&Slam&&\\
\cite{rau2019:IJCASR}&2019&cGAN(pix2pix)&Self&Phantom, Real,&Adversarial&Phantom, Real,&Depth\\
&&&&Synthetic&&Synthetic&\\
\cite{liu2019:TMI}&2019&Siamese+&Self&Real&Monocular&Real&Depth\\
&&DenseNet&&&&&\\
\cite{gomez2021sd}&2020&SD-DefSLAM&Mandala,&Real&Monocular&Real&Depth\\
&&&Hamlyn&&SLAM&&\\
\cite{liu2020depth:MICCAI}&2020&Siamese+&Private&Real&Monocular& Real&Depth,\\
&&DenseNet&&&self-supervised&&3D reconstruction\\
\cite{xi2021:CMPBM}&2021&DepthNet+&EndoAbs, Self&Real,&Monocular&Real,&Depth,\\
&&ConfidenceNet&Private&Synthetic&&Synthetic&3D point cloud\\
\cite{li2021:ICRA}&2021&U-Net+&Self&Phantom&Robot data&Phantom&Depth,\\
&&Robodepth&&&&&FOV Control\\
\cite{cheng2021:MICCAI}&2021&DepthNet&UCL&Synthetic,&Adversarial&Synthetic,&Depth\\
&&&&Real&&Real&\\
\cite{long2021:MICCAI}&2021&E-DSSR&Hamlyn,&Real&Stereo&Real&Depth\\
&&&Private&&&&\\
\cite{huang2021:MICCAI}&2021&SADepth&dVPN,&Real&Adversarial,&Real&Depth\\
&&&SCARED&&Stereo&&\\
\cite{yang2021dense:MIUA}&2021&END-Flow&SCARED&Real&Stereo&Real&Depth\\
&&&&&unsupervised&&\\
\cite{lu2021super:ICRA}&2021&DNN+&SuPer,&Real&Stereo&Real&Depth,\\
&&DeepLabCut&Hamyln&&&& Tissue Tracking\\
\cite{recasens2021endo}&2021&DNN&Hamyln&Real&Monocular&Real&3D Reconstruction\\
&&&&&&&Camera pose\\
\bottomrule
\end{tabular}
\begin{tablenotes}
$^{\rm a}$ Modified Model ;                      
\end{tablenotes}
\end{threeparttable}
\end{center}
\end{minipage}
\end{sidewaystable*}


\subsubsection{Stereo depth techniques}
\begin{figure*}[t!]
\captionsetup{justification=justified}
\centering
\includegraphics[width=\textwidth,height=3in, keepaspectratio]{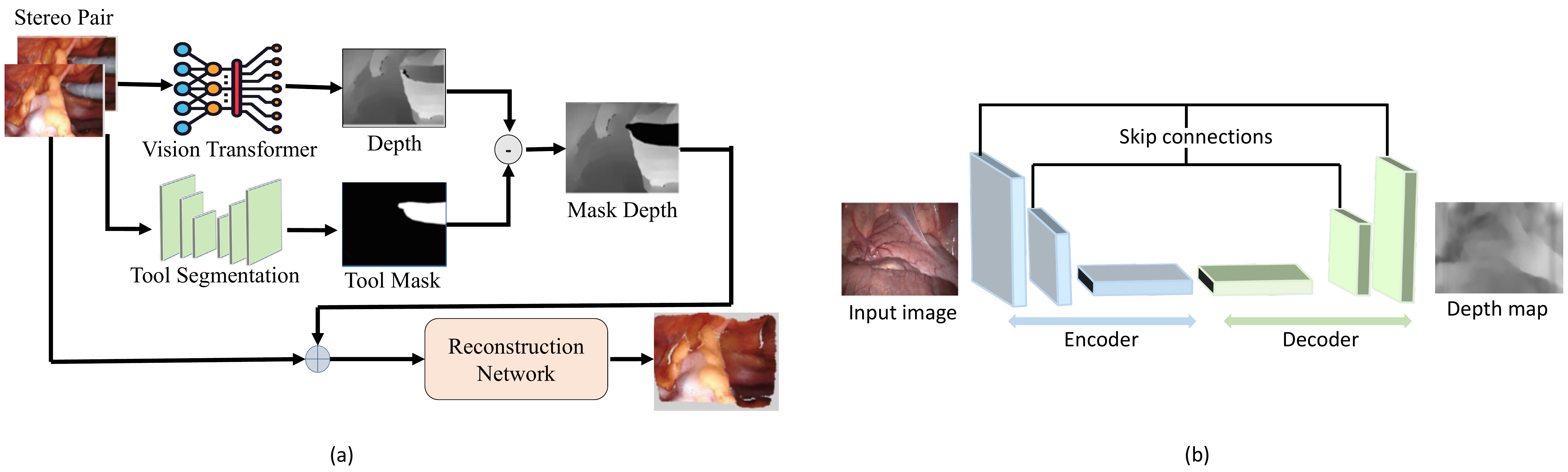}
\includegraphics[width=\textwidth,height=3in, keepaspectratio]{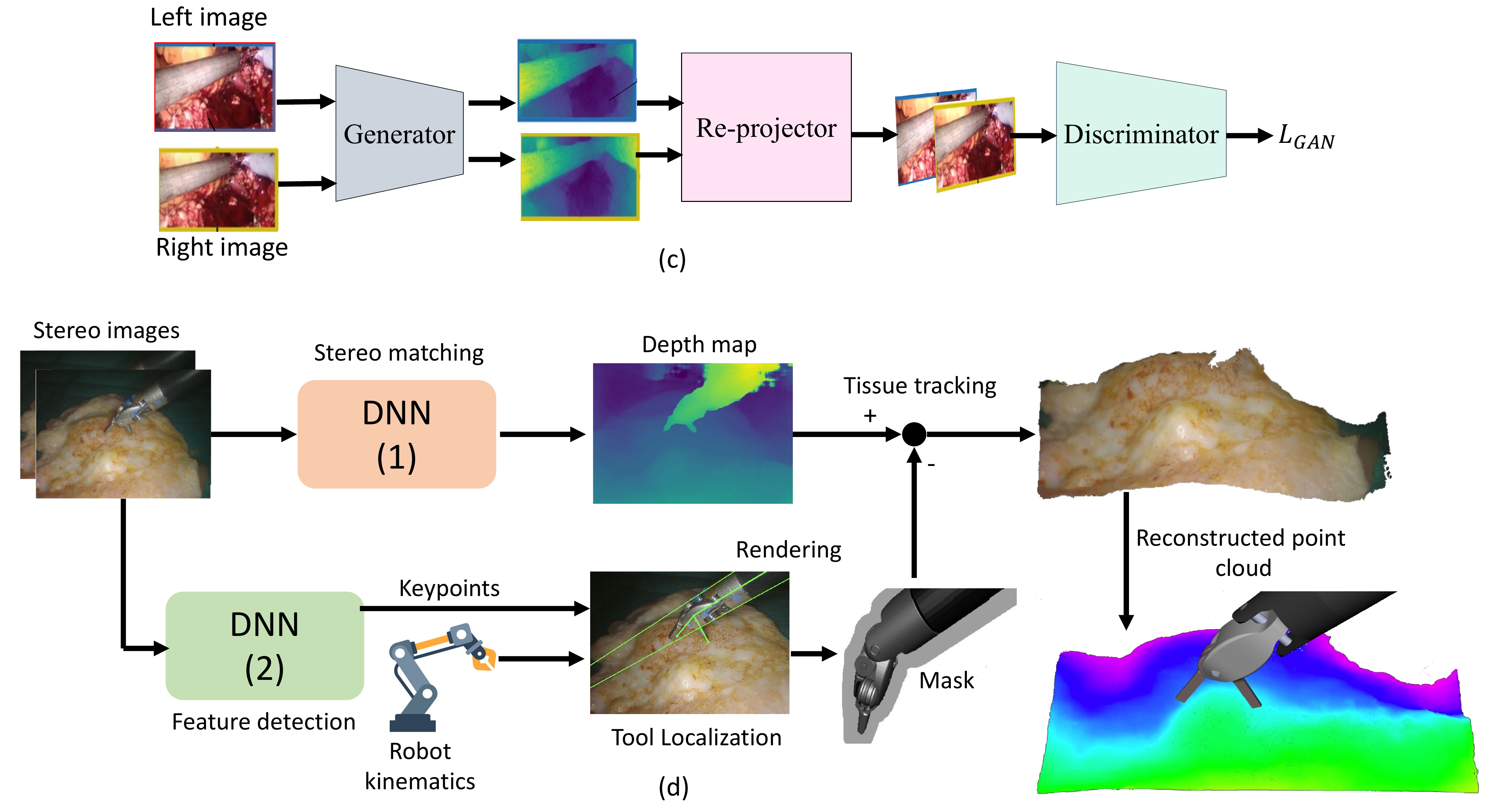}
\caption{Depth Perception and 3D reconstruction architectures. a) E-DSSR \citep{long2021:MICCAI} uses a stereoscopic transformer for depth estimation while a lightweight segmentation  odel to tackle tool occlusions, b) DepthNet \citep{xi2021:CMPBM} takes monocular image and computes depth. The network contains 14 layers of an encoder and as many decoder layers, b)SA-Depth \citep{huang2021:MICCAI} is a self Supervised Depth estimation framework, which uses UNet like architecture in the generator network, while the discriminator contains CNN, BatchNorm and activation function layers,  d) Super Deep network \citep{lu2021super:ICRA} uses DNN(1) to compute depth by stereo matching technique. The DNN(2) uses keypoints alonwith robot kinematic information to estimate tool pose. Final scene is reconstructed by fusing deformable tissue point cloud with depth maps.} 
\label{fig7}
\end{figure*}

Depth estimation from monocular images is quite challenging especially in endoscopic video since obtaining camera pose parameters is difficult. This has led researchers to work in stereo endoscopes. In stereo endoscopy, depth estimation is performed using pixel matching between two binocular pairs. Afterwards, the matched points can be triangulated to recover the depth map. Two types of approaches in this direction have been studied- traditional and deep learning based. Traditional methods usually use optical flow \citep{phan2019:ICIP} or stereo-matching \citep{geiger2010:ACCV} techniques. Several studies have been conducted that use stereo-matching algorithms to estimate endoscopic scene depth. For instance,  a light-weight network for real-time depth estimation is proposed in \citep{gan2021:Neuro} that uses pseudo-convolutions instead of normal convolutions. The proposed network tries to balance the trade-off between accuracy and inference time. Despite achieving good results though, depth estimation methods that use triangulation have certain domain-specific limitations like the presence of texture-less surfaces, occlusions, and specular reflections to name a few.  In another work \citep{li2021:CVF}, authors use Efficient Large-Scale Stereo (ELAS) matching to generate a depth map from stereo images. However, the depth maps generated undesirable noise and the approach uses colors for tool tracking which makes it susceptible to illumination changes. Authors in \cite{lu2021super:ICRA} propose a dual DNN framework (SuPer Deep) for tool and tissue tracking. One uses stereo image information for depth estimation and subsequently tissue tracking while other DNNs uses kinematic data to do tool tracking. Depth estimation in endoscopy data becomes quite challenging in the presence of tissue deformations and occlusions. To this end, E-DSSR \citep{long2021:MICCAI} proposes transformer based reconstructions architecture for a dynamic surgical scene (see Fig.\ref{fig7}(b)). Results show promising performance in terms of SSIM and PSNR and it is fourteen times faster than the previous method \citep{li2021:CVF}. 

Since, it is difficult to obtain depth labels, self-supervised approaches have also been studied to train models with stereo-endoscopic data. In this regards, Siamese learning with spatial transformer and auto-encoder is proposed in \cite{ye2017:arXiv} to estimate depth of surgical scenes. A GAN based architecture named SADepth \citep{huang2021:MICCAI} was proposed for surgical scene depth estimation using a pair of stereo images. The SADepth generator takes in left-right image pair and produces a depth disparity map, while reproduction sampler \citep{jaderberg2015:NIPS} reconstructs camera input images and then reconstructed and original images are fed to the discriminator to distinguish between real and fake (see Fig.\ref{fig7}(a). The results indicate that the model performs better than SOTA in SSIM evaluation measure. 

Training a self-supervised framework requires stereo image pairs to be rectified first. It may be easier to do in the case of natural images, but in an endoscopic domain, images are texture-less and hard to find matching points thereby making the training process quite tedious and prone to rectification errors. Unsupervised optical flow-based methods, on the other hand, do not require stereo-rectified images. To this end, an unsupervised depth estimation method named END-Flow \citep{yang2021dense:MIUA} is proposed without needing ground truth labels, rectified images or camera calibration parameters. The results show competitive performance on SCARED dataset\citep{allan2021stereo}.

\subsubsection{SLAM-based methods} 
Visual Simultaneous localization and mapping (Visual SLAM) has been employed in a number of depth perception and 3D reconstruction applications. The main advantage of SLAM is that it can fulfill real-time requirements of the system by providing quick feedback about the endoscope location  concerning human tissues or depth information of internal organs to the surgeon. In one of the initial works in surgical navigation, 3D reconstruction of the abdominal cavity using visual SLAM was investigated from monocular endoscopic data in \cite{grasa2013:TMI}. Other SLAM-based approaches \citep{marmol2019:IEEE,mahmoud2018:IEEE} have used features of stereo images for depth estimation. However, the lack of discernible features in surgical images and lighting variations makes it difficult to implement feature-based methods. To that end, \cite{chen2019slam:arXiv} proposed an adversarial approach to estimate depth from monocular images and then 3D reconstruction of the surgical scenes by using SLAM technique to fuse the original image and the predicted depth. In another work \citep{qiu2018:CVPR}, ORB-SLAM \citep{mur2015orb:TR} was used with laser light markers as artificial feature points for dense map generation of the oral cavity. Taking the initial sparse reconstruction as input from abdominal exploration, authors in \cite{mahmoud2017:arXiv} propose a quasi dense 3D reconstruction algorithm for MIS. Extending this work further, \cite{mahmoud2018:TMI} modify ORB-SLAM and use a few keyframes only for dense reconstruction. Results indicate that system could not perform well in cases of soft texture-less tissues or high tissue deformations. In an effort to model highly deformable scenes, SD-DefSLAM \citep{gomez2021sd} was proposed with an illumination-invariant optical flow method and ORB feature extraction. The model was applied to challenging images containing specular reflections, occlusions, and weak texture scenarios. The results indicate that model can address these domain-specific challenges in an efficient way.  

\section{Benchmark datasets}\label{sec6}

Learning-based methods are always reliant on the excess amount of training data and efficiency is usually proportional to the number of available training examples. This makes the availability of datasets an important prerequiste for a robust deep learning model. A ummary of datasets used in surgical tool navigation is provided in Table \ref{tab8}. It is pertinent to mention here that MICCAI EndoVis challenges contain several sub-challenges, details of only those are provided which are relevant with this study. Most of the datasets in this domain are publicly available which serve two purposes-researchers who can not obtain their own data can use them and contribute towards building models and they can be used for comparing various frameworks developed in the literature. Looking at the datasets, it can be inferred that most of them provide video data except for one which also provides kinematic data. With the rise of robotic surgery, it is anticipated that video datasets will be accompanied by robot kinematics too to make more robust architectures. 

A significant contribution towards the advancement of MIS and the integration of data-driven techniques into interventional healthcare has been accomplished by the Medical Image Computing and Computer Assisted Intervention (MICCAI)\footnotemark[1] society. MICCAI hosts annual challenges on various aspects of the surgical domain annually since 2015 which has greatly advanced AI-based research and practice in computer vision applications into the computer-assisted interventions. A brief description of each dataset is provided below.

\noindent \textbf{EndoVis challenge}: The Endoscopic vision challenges have been organized by the MICCAI society every year under different themes, starting from 2015. The ENdoVis datasets contain both rigid and robotic instruments along with tool annotations. The first challenge focused on instrument segmentation and tracking for both rigid and articulated tools. The second challenge (from 2017) contained robotic-tool videos for binary and multi-class instrument segmentation. The task complexity of surgical videos was extended to include whole surgical scene segmentation in 2018. In this challenge, participants were required to segment robotic, non-robotic and anatomical objects in the surgical video frames. Robustness and generalization was the theme of 2019 Robust-MIS 2019 challenge in which surgical video dataset was collected from thirty different procedures. The challenge objectives were instrument binary segmentation, parts and instance segmentation. In the same year, another sub-challenge, the Stereo Correspondence and Reconstruction of Endoscopic Data (SCARED) focused on depth estimation from surgical scenes. In 2020, the MICCAI sub-challenge covered surgical visual domain adaptation, while in 2021 addressed the workflow recognition problem.

\footnotetext[1]{For details, refer to \textbf{https://endovis.grand-challenge.org/}}

\noindent \textbf{Cholec80}: This dataset \citep{endonet:TMI} contains 80 videos of cholecystectomy procedures performed by 13 clinicians. The dataset contains tool presence and phase recognition annotations. A tool is labelled as to be present if at least half of the tool tip is visible in the video frame. Video capture speed is 25 fps and is downsampled to 1 fps for processing. Cholec80 is split into two subsets, each containing 40 videos. The first subset, known as the fine tuning subset contains 86K annotated images with 10 videos annotated with tool bounding boxes. The second subset is named evaluation subset which is used to evaluate models for tool presence detection and phase recognition. 
    
Cholec80 has been further extended by adding 40 additional annotated cholecystectomy videos \citep{aksamentov2017:springer}. In another extension, ITEC Smoke\_Cholec80 Image contains 100K frames from cholec80 essentially for smoke removal. The target classes are smoke and non-smoke. 

\noindent \textbf{M2CAI16:} This challenge is comprised of two datasets for two different tasks, surgical workflow and tool detection. The m2cai-tool\footnotemark[1] dataset \citep{endonet:TMI} contains 15 cholecystectomy videos (10 for training the model and 5 for test) collected in collaboration with University Hospital of Strasbourg. The task was to identify the instruments present in the surgical scene. Surgical videos had been annotated with binary tool presence with 7 tools in total. m2cai16-tool dataset has been extended to include spatial annotations for 2,532 frames across the first 10 videos of the 15 total 15 videos. This dataset is named m2cai-tool-locations \citep{jin2018tool:WACV}. With this extension, it is easier to localize surgical tools in addition to performing tool classification. 

\noindent \textbf{ATLAS Dione:} This dataset contains 99 surgical videos of 6 different tasks performed by 10 clinicians at Roswell Park Cancer Institute (Buffalo, NY) \citep{sarikaya2017:TMI}. The surgical procedures were performed on \emph{da Vinci} Surgical System. The tasks comprise basic robotic surgery skills named Fundamental Skills of Robotic Surgery (FSRS) and special skills for Robotic Anastomosis Competency Evaluation (RACE). The annotations are provided in terms of tool bounding boxes, surgical actions, duration and surgeon skills levels. 
\noindent \textbf{UCL dVRK dataset:} This dataset \cite{colleoni2020synthetic:MICCAI} consists of 14 videos of 300 frames each having the frame size of 720x576 recorded using \emph{da Vinci Research Kit}. Data collection involved 5 different kinds of animal tissues (chicken breast and back, lamb and pork loin, beef sirloin) with varying backgrounds and illumination conditions. To make the test set more challenging, lamb kidneys and blood were placed in the background. Fractional Brownian Motion was also added in the test frames to induce noise. Surgical tool annotations are recorded in terms of segmentation masks. This dataset also provides robot kinematic information.

\noindent \textbf{LapSig300:} LapSig300 \citep{kitaguchi2020:IJS} is a large collection of data comprising 300 videos of laparoscopic colorectal surgery. The dataset was assembled in collaboration with 19 high-volume institutions of japan. There are a total 82, 623, 098 frames in the dataset which are annotated for surgical phase and action recognition while 4243 frames are annotated for semantic segmentation of tools. Five tools are selected for segmentation based on their frequency of occurrence in the data, grasper, dissector, linear dissector, Maryland, and clipper. The dataset is available for use at the author's request. 
\footnotetext[1]{http:
//camma.u-strasbg.fr/m2cai2016/index.php/
tool-presence-detection-challenge-results.}

\noindent \textbf{NeuroSurgicalTools:} This surgical tool detection dataset \citep{bouget2015detecting:TMI} is composed of 14 monocular videos assembled through "Zeiss OPMI Pentero classic" microscopes. The video capture parameters are 720×576 pixels at 25 fps during in-vivo neurosurgery performed at CHU Pontchaillou, Rennes. The dataset contains multiple tool challenge scenarios such as tool occlusions by organs or surgeon hands, tools overlapping, tools covered by blood, blurriness, and specular reflections. Seven different surgical instruments are featured in the dataset. Every tool is annotated with a bounding polygon in addition to each part of the tool is given a multi-class label. Tool orientation, its width, and tool tip information is also made part of the dataset. 
\noindent \textbf{FetalFlexTool:} This is an ex-vivo fetal surgery dataset \citep{garcia2016real} consisting of 21 images for model training and one 10-second video for testing. Non-rigid McKibben artificial muscle actuation was used to record the dataset. All the training images were captured in the air while the video was recorded underwater to have varying lighting conditions and backgrounds. Tool segmentation masks were manually annotated.

\noindent \textbf{LapGyn4\footnotemark[1]:} Comprising over 55K images, LapGyn4 is a four-part Gynecological surgery dataset incorporating data from 500 interventions. It comprises surgical scenes relating to surgical actions, anatomical structures, visible surgical tools, and actions performed on particular anatomy.

\footnotetext[1]{https://zenodo.org/record/1219280\#.X6E4O4hKiUk}

\noindent \textbf{dVPN Dataset:} The in-vivo dVPN dataset \citep{ye2017:arXiv} is collected from \emph{da Vinci} partial nephrectomy procedure. It contains 34320 pairs of rectified stereo images for training and 14382 pairs for testing. No ground truth labels are available for the dataset.

\noindent \textbf{UCL:} UCL \citep{rau2019:IJCASR} is a synthetic dataset generated from a human CT colonography (CTC) scan. Manual segmentation and meshing is used to extract surface mesh. In order to render endoscopic images with their depth information, Unity game engine application is used. A virtual camera with two light sources run through the virtual model producing different images containing various illumination scenarios. The virtual materials contain various textures to make the dataset more versatile. Overall, the dataset consists of  16,000 images along with as many depth maps.  

\begin{sidewaystable*}[!htbp]
\tabcolsep7.5pt
\footnotesize
\begin{minipage}{\textheight}
\caption{Tool Navigation Datasets}\label{tab8}
\begin{center}
\begin{threeparttable}
\begin{tabular}{@{}l|l|l|l|l|l|l|l@{}}
\toprule

\textbf{Dataset} & \textbf{Year} &  \textbf{Data Size}  & \textbf{Procedure} &   \textbf{License} & \textbf{Tools}&  \textbf{Annotations}& \textbf{Tasks} \\

\hline

EndoVis 15&2015&9K images&colorectal surgery&Public&Rigid&Pixel-wise,2D pose&Seg$^{\rm a}$, T$^{\rm a}$\\
&&&&&Robotic&&\\
NeuroSurgicalTools&2015&2476 images&Neurosurgery&Public&Rigid&Bounding-box&D$^{\rm a}$\\
FetalFlexTool&2015&21 images&Fetal Surgery&Public&Rigid&Bounding-box&D$^{\rm a}$\\
&&One video&&&&\\
M2CAI16-tool&2016&16 Videos&cholecystectomy&Public&Rigid&TP$^{\rm h}$&D$^{\rm a}$\\
Cholec80&2016&80 Videos&cholecystectomy&Public&Rigid&TP$^{\rm h}$, Phase&D,PR$ {\rm a}$\\
EndoVis 17$^{\rm b}$&2017&10 Videos&Porcine&Public&Robotic&Piwel-wise&Seg(B,P,I$^{\rm c}$)\\
ATLAS Dione&2017&86 Videos&In-vitro Experiments&Public&Robotic&Bounding Box&D,L$^{\rm a}$,Activity, Skills\\
Hamyln&2017&2 Phantom&Cardiac&Public&NA&Depth map&TT$^{\rm a}$\\
EndoVis 18$^{\rm d}$&2018&14 Videos&Nephrectomy&Public&Robotic&Pixel-wise mask&Scene Seg$^{\rm a}$\\
LapGyn4&2018&55K images&Gynecologic Surgery&Public&Rigid& No annotation&Multiple$^{\rm e}$\\
m2ccai16-tool&2018&16 Videos&cholecystectomy&Public&Rigid&Bounding-box&D$^{\rm a}$\\
locations&&&&&&&\\
ROBUST-MIS19&2019& 30 videos&proctocolectomy&Public&Rigid&Instances&Seg(B,P,I$^{\rm d}$)\\
&&&rectal resection&&&&\\
&&&sigmoid resection$^{\rm *}$&&&&\\
UCL&2019&16016 Synthetic&colonoscopy&Public&NA&Depth map&DE$^{\rm a}$\\
&&images&&&&&\\
Cata7&2019&7 Videos&Cataract Surgery&Private&Rigid&Pixel-wise mask&Seg(I$^{\rm d}$)\\
SCARED$^{\rm f}$&2019&27 Videos&Porcine&Public&NA&Depth+&3D reconstruction\\
&&&&&&camera parameters&\\
UCL dVRK&2020&20 Videos+&Ex-Vivo&Public&Robotic&Pixel-wise&Seg(B$^{\rm d}$)\\
&&Kinematic Data&&&&&\\
Sinus Surgery-C&2020&10Videos&Sinus-Cadaver&Public&Rigid&Pixel-wise mask&Seg(B$^{\rm c}$)\\
Sinus Surgery-L&2020&3 Videos&Sinus-Live&Public&Rigid&Pixel-wise mask&Seg(B$^{\rm c}$)\\
LapSig300&2020&300 Videos&Colorectal Surgery&Private&Rigid&Pixel-wise mask&Seg(I$^{\rm c}$), PR,AR$^{\rm a}$\\
&&&&&&Phase, Action&\\
EndoVis 21$^{\rm g}$&2021&33 Videos&cholecystectomy&Public&Rigid&TP, A,SC,Ph$^{\rm h}$&D,PR,AR$^{\rm a}$\\
dVPN&2021&48702 images&nephrectomy&Private&NA&NA&DE$^{\rm a}$\\
CaDTD&2021&50 Videos&Cataract&Public&Rigid&Bounding-box&D$^{\rm a}$\\
&&&Surgery&&&&\\
SCARED&2021&9 datasets&Porcine&Public&Robotic&depth&DE$^{\rm a}$\\
&&4-5 keyframes each&&&&&\\
\hline
\end{tabular}
\begin{tablenotes}
$^{\rm a}$Seg=Segmentation, T=Tracking, D=Detection PR=Phase Recognition, L=Localization, DE=Depth estimation TT=Tissue Tracking AR=Action Recognition;  $^{\rm b}$Robotic Instrument Segmentation Sub-Challenge; $^{\rm c}$ B=Binary, P=Parts detection, I=Instance Segmentation; $^{\rm d}$ Robotic Scene Segmentation Sub-Challenge;$^{\rm e}$instrument counts, action detection, anatomical structures; $^{\rm f}$ Stereo Correspondence and Reconstruction of Endoscopic Data Sub-Challenge; $^{\rm g}$ Surgical Workflow and Skill Analysis;$^{\rm h}$ TP=Tool presence, A=Action, SC=Skill Classification, Ph=Phases; 
$^{\rm *}$ Unkmown Surgery
\end{tablenotes}
\end{threeparttable}
\end{center}
\end{minipage}
\end{sidewaystable*}

\noindent \textbf{Cata7:} This is the first cataract surgery dataset \citep{ni2019raunet} recorded at Beijing Tongren Hospital, containing 7 videos and each video featuring a full cataract surgery procedure. The videos are spilt into images of resolution of 1920×1080 pixels. Videos are downsampled from 25 fps to 1 fps to avoid redundancy. Data annotations are provided for surgical instrument types and with precise edges.

\noindent\textbf{Sinus Surgery Dataset:} The sinus surgery datasets \citep{qin:IEEE} are composed of two cadavers and live data portions. The Cadaver dataset consists of 10 cadaver sinus surgery videos performed on 5 cadaver subjects and involved 9 surgeons. Each subject was operated on both right and left nasal cavities. The dataset's video duration ranges from 5 minutes to 23 minutes with a resolution of 320×240 at 30 fps.  The live dataset comprises 3 videos whose duration ranges from 12 minutes to 66 minutes with a resolution of 1920×1080. This dataset contains various challenging scenes such as blurry frames, smoke, instruments in shadow, tissue occlusions, and specular reflections. All the video frames in the dataset were center-cropped with 240×240 size and manually annotated with foreground surgical tools.

\noindent\textbf{CaDTD:} This is a cataract surgery dataset \citep{jiang:Springer} containing 50 videos taken from CATARACTS dataset \citep{al2019:MIA}. The surgeries were performed by renowned clinicians at Brest University Hospital. Videos were captured at 30 fps and having a resolution of 1920×1080. Half of the videos are unlabelled while half labelled with tool bounding box annotations. To make the videos more usable, they were downsampled to a resolution of 720 × 540 with every frame being 3s apart. The dataset defines tools into two different configurations. One category assumes a whole tool including its head and handles while the other only considers tool heads. 

\noindent \textbf{SCARED}: SCARED dataset \cite{allan2021stereo} was released as part of Enndovis sub-challenge Stereo Correspondence and Reconstruction of Endoscopic Data (SCARED) at MICCAI 2019. It contains 7 training and 2 test datasets captured through \emph{Da Vinci Xi} surgical robot. Each dataset contains structured light data of a single porcine subject. All the keyframes of data also contain depth labels.

\section{Current gaps and future directions}\label{sec7}
We presented a comprehensive overview of methods developed in surgical tool navigation and depth perception. Afterward, we described benchmark datasets available in the surgical domain. The literature on surgical AI points towards the increasing interest of the research community which has led towards developing robust models, there are still some research gaps and important questions that need to be addressed. Those issues are elaborated upon in this section. 

\noindent \textbf{1) Data-related gaps}: One of the important hurdles in integrating AI into surgery is the availability of sufficient annotated training data.  
The performance of AI-based models relies heavily upon data availability. Also, the data labeling requires expert annotators. \cite{maier2022surgical} argues that the lack of success stories in surgery as contrasted to other medical domains, such as radiology and dermatology can be attributed to the lack of quality in the annotated datasets. The authors further mentioned EndoVis \cite{EndoVis17:arXiv}, Cholec80 \cite{endonet:TMI} and JIGSAW \cite{jigsaw} as the most notable datasets in the surgical domain, but the small size and limited diversity are still needed to be addressed. Another concern in the surgical domain is the variety of surgeries and rapid changes in surgical techniques, which might render datasets obsolete with time- a problem not observed in traditional image domains. 

In this context, quality assurance of data annotation needs to be established to ensure the model performance and its reliability. Furthermore, given that the size and variety of existing datasets is limited, a valuable initiative can be to develop new large and diverse surgical datasets covering various surgical domain tasks. Another way forward to overcome data size, accessibility, data privacy, and ethical concerns can be exploiting federated learning.

\noindent \textbf{2) Current Method Development Trends}
Several methods have been developed to tackle surgical navigation problems including self-supervised methods to address the lack of annotations \cite{teevno2022}, attention mechanisms to reduce the impact of artifacts in tool recognition \citep{ni:EMBC}, and adversarial approaches to generate synthetic data for model robustness \cite{colleoni2022ssis}. Existing methods have produced promising results but more needs to be done. For instance, the methods developed for surgical tool localization and segmentation may be improved by considering inter-frame semantic relationships to improve segmentation performance. The use of augmented reality (AR) also remains quite under-explored in the literature. We suggest that the integration of AR into surgery can be extremely helpful to the surgeon to better understand surgical scenes. 

For clinical usefulness, algorithms need to have faster inference times, and need to be tested on sufficiently fast hardware to enable real-time execution. If additional information such as pre-operative data is required, algorithms need to be able to access that data. These problems can be addressed by developing appropriate infrastructure or creating test environments such as experimental operating rooms where real time algorithms can be validated and evaluated in a realistic manner.

\noindent \textbf{3) Adaptation to Clinical Procedures in Action}
Clinical translation of AI-based methods in surgery has been prohibitively slowed down due to several factors. First, long term clinical studies which are an important prerequisite for a clinical translation, despite intense research work is in its early stages. Secondly, lack of standardized surgical procedures impede the creation of a standard data annotation protocol, which is essential for multi-center studies. Thirdly, digitization has not yet been fully established in the operating room and surgical community compared to other medical domains. Therefore, a possible way forward may be to focus on factors such as clinical usefulness, easy workflow integration, technical viability and high business value to attract industry participation. 

\noindent \textbf{4) Robot-assisted Surgery}
Surgical robots are being used successfully in major surgeries such as esophagus, pancreas and rectum, but their actual share in overall procedures is still marginal \cite{gumbs2021artificial}. Several attempts of controlled trials have not yielded fruitful results and conventional laparoscopic surgery even surpassed robot-assisted surgery (RAS) in terms of operative time and cost and total complication rate \cite{roh2018robot}. A possible way forward for RAS is to focus on reducing operative cost and time, blood loss, complication rate and length of hospital stay. 

Modern robots such as \emph{da Vinci} surgical system (Intuitive Surgical Inc., Sunnyvale, CA, USA) and its other competitors can add value to the usefulness in enhancement and automation of surgical procedures. However, correcting the camera positioning by performing calibrations techniques already demonstrates an example of tedious task for the surgeon. Developing an automatic camera positioning system could be a good starting point for enhanced use of surgical robots in the operating room.

\section{Conclusion}\label{sec8}
Advancements in deep learning methods development has accelerated the pace of surgical data science research. More and more methods for surgical instrument navigation are being investigated paving a way for a better surgical care, patient safety, and reducing surgeon's visual load during MIS procedures. The volume of research in the last five years shows that the instrument segmentation and detection have received most of the attention of researchers followed by tracking and scene depth estimation. Improving model robustness towards artifacts in the endoscopic data and reducing the annotation burden have been the most researched areas in the surgical navigation literature. Despite promising outcomes, several research gaps related to data availability and annotation, method development, clinical adaptation of those methods and robot-assisted surgery still persist. Addressing these gaps can help in assisting surgeons in various surgical phases and improve patient outcomes. \\

\noindent\textbf{Conflict of interest}:

The authors do not have any conflicts of interest to disclose.

\end{sloppypar}

\bibliography{sn-bibliography}



\end{document}